\documentclass{article} 
\usepackage{iclr2020_conference_arxiv,times}

\usepackage{amsmath,amsfonts,bm}

\def\eqref#1{equation~\ref{#1}}

\def\1{\bm{1}}

\def\rb{{\textnormal{b}}}
\def\rc{{\textnormal{c}}}

\def\re{{\textnormal{e}}}

\def\rp{{\textnormal{p}}}

\DeclareMathAlphabet{\mathsfit}{\encodingdefault}{\sfdefault}{m}{sl}
\SetMathAlphabet{\mathsfit}{bold}{\encodingdefault}{\sfdefault}{bx}{n}

\newcommand{\sigmoid}{\sigma}

\DeclareMathOperator*{\argmax}{arg\,max}
\DeclareMathOperator*{\argmin}{arg\,min}

\usepackage[hyphens]{url}
\usepackage{hyperref}

\usepackage{microtype}
\usepackage{graphicx}
\usepackage{subcaption}
\usepackage{booktabs} 

\usepackage{amsmath}
\usepackage{amsfonts}
\usepackage{amsthm}

\title{V-MPO: On-Policy Maximum a Posteriori\\ Policy Optimization for Discrete and\\ Continuous Control}

\author{H. Francis Song\thanks{Equal contribution}, Abbas Abdolmaleki\footnotemark[1], Jost Tobias Springenberg, Aidan Clark, \\ {\bf Hubert Soyer, Jack W. Rae, Seb Noury, Arun Ahuja, Siqi Liu, Dhruva Tirumala}, \\ {\bf Nicolas Heess, Dan Belov, Martin Riedmiller, Matthew M. Botvinick} \\ DeepMind, London, UK \\ \texttt{\{songf,aabdolmaleki,springenberg,aidanclark,}\\ \texttt{ soyer,jwrae,snoury,arahuja,liusiqi,dhruvat,}\\ \texttt{ heess,danbelov,riedmiller,botvinick\}@google.com}}

\iclrfinalcopy 
\begin{document}

\maketitle


\begin{abstract}
Some of the most successful applications of deep reinforcement learning to challenging domains in discrete and continuous control have used policy gradient methods in the on-policy setting. However, policy gradients can suffer from large variance that may limit performance, and in practice require carefully tuned entropy regularization to prevent policy collapse. As an alternative to policy gradient algorithms, we introduce V-MPO, an on-policy adaptation of Maximum a Posteriori Policy Optimization (MPO) that performs policy iteration based on a learned state-value function. We show that V-MPO surpasses previously reported scores for both the Atari-57 and DMLab-30 benchmark suites in the multi-task setting, and does so reliably without importance weighting, entropy regularization, or population-based tuning of hyperparameters. On individual DMLab and Atari levels, the proposed algorithm can achieve scores that are substantially higher than has previously been reported. V-MPO is also applicable to problems with high-dimensional, continuous action spaces, which we demonstrate in the context of learning to control simulated humanoids with 22 degrees of freedom from full state observations and 56 degrees of freedom from pixel observations, as well as example OpenAI Gym tasks where V-MPO achieves substantially higher asymptotic scores than previously reported.
\end{abstract}


\renewcommand{\d}{\partial}
\newcommand{\der}[2]{\frac{d#1}{d#2}}
\newcommand{\ders}[2]{\frac{d^2#1}{d#2^2}}
\newcommand{\pder}[2]{\frac{\partial #1}{\partial #2}}
\newcommand{\pders}[2]{\frac{\partial^2 #1}{\partial #2^2}}
\newcommand{\pdersmixed}[3]{\frac{\partial^2 #1}{\partial #2 \partial #3}}
\newcommand{\br}[1]{\underline{#1}}
\newcommand{\lp}{\bigg(}
\renewcommand{\rp}{\bigg)}
\newcommand{\Lp}{\Bigg(}
\newcommand{\Rp}{\Bigg)}
\newcommand{\lb}{\bigg[}
\renewcommand{\rb}{\bigg]}
\newcommand{\Lb}{\Bigg[}
\newcommand{\Rb}{\Bigg]}
\newcommand{\lc}{\bigg\{}
\renewcommand{\rc}{\bigg\}}
\newcommand{\Lc}{\Bigg\{}
\newcommand{\Rc}{\Bigg\}}
\newcommand{\pf}[2]{\lp\frac{#1}{#2}\rp}
\newcommand{\sqrtfrac}[2]{\sqrt{\frac{#1}{#2}}}

\let\oldvec\vec
\renewcommand{\vec}[1]{\mathbf{#1}}

\newcommand{\ket}[1]{|#1\rangle}
\newcommand{\bra}[1]{\langle#1|}
\newcommand{\braket}[2]{\langle#1|#2\rangle}
\newcommand{\mean}[1]{\langle#1\rangle}
\newcommand{\Mean}[1]{\big\langle #1 \big\rangle}
\newcommand{\tr}{\operatorname{Tr}}
\newcommand{\colv}[2]{\left(\begin{matrix} #1 \\ #2 \end{matrix}\right)}
\newcommand{\rowv}[2]{\left(\begin{matrix} #1 & #2 \end{matrix}\right)}
\newcommand{\mat}[4]{\left(\begin{matrix} #1 & #2 \\ #3 & #4 \end{matrix}\right)}
\newcommand{\sinc}{\operatorname{sinc}}
\newcommand{\sech}{\operatorname{sech}}
\newcommand{\Si}{\operatorname{Si}}
\renewcommand{\re}{\operatorname{Re}}
\newcommand{\im}{\operatorname{Im}}
\newcommand{\erf}{\operatorname{erf}}
\newcommand{\erfc}{\operatorname{erfc}}
\newcommand{\erfi}{\operatorname{erfi}}
\newcommand{\sgn}{\operatorname{sgn}}

\renewcommand{\sigmoid}{\operatorname{sigmoid}}

\newcommand{\be}{\begin{equation}}
\newcommand{\ee}{\end{equation}}

\newcommand{\unit}[1]{\text{ #1}}

\newcommand{\var}{\operatorname{Var}}
\newcommand{\cov}{\operatorname{Cov}}

\newcommand{\maxover}{\operatornamewithlimits{max}}

\renewcommand{\argmax}[1]{\operatorname{arg}\operatornamewithlimits{max}_{#1}}
\renewcommand{\argmin}[1]{\operatorname{arg}\operatornamewithlimits{min}_{#1}}

\newcommand{\stopgradient}{\operatorname{sg}}

\newcommand{\dkl}{D_\text{KL}}
\newcommand{\expectation}[1]{\operatornamewithlimits{\mathbb{E}}_{#1}}


\renewcommand{\cite}[1]{\citep{#1}}



\newcommand{\figschematic}{\ref{fig:schematic}}

\section{Introduction}
\label{introduction}

Deep reinforcement learning (RL) with neural network function approximators has achieved superhuman performance in several challenging domains~\cite{Mnih2015,Silver2016,Silver2018}. Some of the most successful recent applications of deep RL to difficult environments such as Dota~2~\cite{OpenAI2018}, Capture the Flag~\cite{Jaderberg2019}, Starcraft~II~\cite{DeepMind2019}, and dexterous object manipulation~\cite{OpenAI2018a} have used policy gradient-based methods such as Proximal Policy Optimization (PPO)~\cite{Schulman2017} and the Importance-Weighted Actor-Learner Architecture (IMPALA)~\cite{Espeholt2018}, both in the approximately on-policy setting.

Policy gradients, however, can suffer from large variance that may limit performance, especially for high-dimensional action spaces~\cite{Wu2018}. In practice, moreover, policy gradient methods typically employ carefully tuned entropy regularization in order to prevent policy collapse. As an alternative to policy gradient-based algorithms, in this work we introduce an approximate policy iteration algorithm that adapts Maximum a Posteriori Policy Optimization (MPO)~\cite{Abdolmaleki2018a,Abdolmaleki2018} to the on-policy setting. The modified algorithm, V-MPO, relies on a learned state-value function $V(s)$ instead of the state-action value function used in MPO. Like MPO, rather than directly updating the parameters in the direction of the policy gradient, V-MPO first constructs a target distribution for the policy update subject to a sample-based KL constraint, then calculates the gradient that partially moves the parameters toward that target, again subject to a KL constraint.

As we are particularly interested in scalable RL algorithms that can be applied to multi-task settings where a single agent must perform a wide variety of tasks, we show for the case of discrete actions that the proposed algorithm surpasses previously reported performance in the multi-task setting for both the Atari-57~\cite{Bellemare2012} and DMLab-30~\cite{Beattie2016} benchmark suites, and does so reliably without population-based tuning of hyperparameters~\cite{Jaderberg2017}. For a few individual levels in DMLab and Atari we also show that V-MPO can achieve scores that are substantially higher than has previously been reported, especially in the challenging Ms. Pacman.

V-MPO is also applicable to problems with high-dimensional, continuous action spaces. We demonstrate this in the context of learning to control both a 22-dimensional simulated humanoid from full state observations---where V-MPO reliably achieves higher asymptotic performance than previous algorithms---and a 56-dimensional simulated humanoid from pixel observations~\cite{Tassa2018,Merel2019}. In addition, for several OpenAI Gym tasks~\cite{Brockman2016} we show that V-MPO achieves higher asymptotic performance than has previously been reported.

\section{Background and setting}

We consider the discounted RL setting, where we seek to optimize a policy $\pi$ for a Markov Decision Process described by states $s$, actions $a$, initial state distribution $\rho_0^\text{env}(s_0)$, transition probabilities $\mathcal{P}^\text{env}(s_{t+1}|s_t,a_t)$, reward function $r(s_t,a_t)$, and discount factor $\gamma \in (0, 1)$. In deep RL, the policy $\pi_\theta(a_t|s_t)$, which specifies the probability that the agent takes action $a_t$ in state $s_t$ at time $t$, is described by a neural network with parameters $\theta$. We consider problems where both the states $s$ and actions $a$ may be discrete or continuous. Two functions play a central role in RL: the state-value function $V^\pi(s_t) = \mathbb{E}_{a_t,s_{t+1},a_{t+1},\ldots} \big[ \sum_{k=0}^\infty \gamma^k r(s_{t+k},a_{t+k}) \big]$ and the state-action value function $Q^\pi(s_t, a_t) = \mathbb{E}_{s_{t+1},a_{t+1},\ldots} \big[ \sum_{k=0}^\infty \gamma^k r(s_{t+k},a_{t+k}) \big] = r(s_t, a_t) + \gamma \mathbb{E}_{s_{t+1}} \big[ V^\pi(s_{t+1}) \big]$, where $s_0 \sim \rho_0^\text{env}(s_0)$, $a_t \sim \pi(a_t|s_t)$, and $s_{t+1}\sim \mathcal{P}^\text{env}(s_{t+1}|s_t,a_t)$.

In the usual formulation of the RL problem, the goal is to find a policy $\pi$ that maximizes the expected return given by $J(\pi) = \mathbb{E}_{s_0,a_0,s_1,a_1,\ldots} \big[ \sum_{t=0}^\infty \gamma^t r(s_t,a_t) \big]$. In policy gradient algorithms~\cite{Williams1992,Sutton2000,Mnih2016}, for example, this objective is directly optimized by estimating the gradient of the expected return. An alternative approach to finding optimal policies derives from research that treats RL as a problem in probabilistic inference, including Maximum a Posteriori Policy Optimization (MPO)~\cite{Levine2018,Abdolmaleki2018a,Abdolmaleki2018}. Here our objective is subtly different, namely, given a suitable criterion for what are good actions to take in a certain state, how do we find a policy that achieves this goal?

As was the case for the original MPO algorithm, the following derivation is valid for any such criterion. However, the \emph{policy improvement theorem}~\cite{Sutton1998} tells us that a policy update performed by exact policy iteration, $\pi(s) = \argmax{a} [Q^\pi(s, a) - V^\pi(s)]$, can improve the policy if there is at least one state-action pair with a positive advantage and nonzero probability of visiting the state. Motivated by this classic result, in this work we specifically choose an exponential function of the advantages $A^\pi(s,a)=Q^\pi(s, a) - V^\pi(s)$.

\textit{Notation.} In the following we use $\sum_{s,a}$ to indicate both discrete and continuous sums (i.e., integrals) over states $s$ and actions $a$ depending on the setting. A sum with indices only, such as $\sum_{s,a}$, denotes a sum over all possible states and actions, while $\sum_{s,a\sim \mathcal{D}}$, for example, denotes a sum over sample states and actions from a batch of trajectories (the ``dataset'') $\mathcal{D}$.
\section{Related work}

V-MPO shares many similarities, and thus relevant related work, with the original MPO algorithm~\cite{Abdolmaleki2018a,Abdolmaleki2018}. In particular, the general idea of using KL constraints to limit the size of policy updates is present in both Trust Region Policy Optimization (TRPO; \citeauthor{Schulman2015a}, \citeyear{Schulman2015a}) and Proximal Policy Optimization (PPO)~\cite{Schulman2017}; we note, however, that this corresponds to the E-step constraint in V-MPO. Meanwhile, the introduction of the M-step KL constraint and the use of top-$k$ advantages distinguishes V-MPO from Relative Entropy Policy Search (REPS)~\cite{Peters2008a}. Interestingly, previous attempts to use REPS with neural network function approximators reported very poor performance, being particularly prone to local optima~\cite{Duan2018}. In contrast, we find that the principles of EM-style policy optimization, when combined with appropriate constraints, can reliably train powerful neural networks, including transformers, for RL tasks.

Like V-MPO, Supervised Policy Update (SPU)~\cite{Vuong2018} seeks to exactly solve an optimization problem and fit the parametric policy to this solution. As we argue in Appendix~\ref{appendix:spu}, however, SPU uses this nonparametric distribution quite differently from V-MPO; as a result, the final algorithm is closer to a policy gradient algorithm such as PPO.
\section{Method}

V-MPO is an approximate policy iteration~\cite{Sutton1998} algorithm with a specific prescription for the policy improvement step. In general, policy iteration uses the fact that the true state-value function $V^\pi$ corresponding to policy $\pi$ can be used to obtain an improved policy $\pi'$. Thus we can

\begin{enumerate}
    \item Generate trajectories $\tau$ from an old ``target'' policy $\pi_{\theta_\text{old}}(a|s)$ whose parameters $\theta_\text{old}$ are fixed. To control the amount of data generated by a particular policy, we use a target network which is fixed for $T_\text{target}$ learning steps (Fig.~\ref{fig:schematic}a in the Appendix).
    \item Evaluate the policy $\pi_{\theta_\text{old}}(a|s)$ by learning the value function $V^{\pi_{\theta_\text{old}}}(s)$ from empirical returns and estimating the corresponding advantages $A^{\pi_{\theta_\text{old}}}(s,a)$ for the actions that were taken.
    \item Estimate an improved ``online'' policy $\pi_\theta(a|s)$ based on $A^{\pi_{\theta_\text{old}}}(s,a)$.
\end{enumerate}

The first two steps are standard, and describing V-MPO's approach to step (3) is the essential contribution of this work. At a high level, our strategy is to first construct a nonparametric target distribution for the policy update, then partially move the parametric policy towards this distribution subject to a KL constraint. Ultimately, we use gradient descent to optimize a single, relatively simple loss, which we provide here in complete form in order to ground the derivation of the algorithm.

Consider a batch of data $\mathcal{D}$ consisting of a number of trajectories, with $|\mathcal{D}|$ total state-action samples. Each trajectory consists of an unroll of length $n$ of the form $\tau=\big[ (s_t,a_t,r_{t+1}), \ldots, (s_{t+n-1},a_{t+n-1},r_{t+n}),\ s_{t+n} \big]$ including the bootstrapped state $s_{t+n}$, where $r_{t+1}=r(s_t,a_t)$. The total loss is the sum of a policy evaluation loss and a policy improvement loss,
\begin{equation}
   \mathcal{L}(\phi, \theta, \eta, \alpha) = \mathcal{L}_V(\phi) + \mathcal{L}_\text{V-MPO}(\theta, \eta, \alpha), \label{eq:total_loss}
\end{equation} where $\phi$ are the parameters of the value network, $\theta$ the parameters of the policy network, and $\eta$ and $\alpha$ are Lagrange multipliers. In practice, the policy and value networks share most of their parameters in the form of a shared convolutional network (a ResNet) and recurrent LSTM core, and are optimized together (Fig.~\ref{fig:schematic}b in the Appendix)~\cite{Mnih2016}. We note, however, that the value network parameters $\phi$ are considered fixed for the policy improvement loss, and gradients are not propagated.

The policy evaluation loss for the value function, $\mathcal{L}_V(\phi)$, is the standard regression to $n$-step returns and is given by Eq.~\ref{eq:value_loss} below. The policy improvement loss $\mathcal{L}_\text{V-MPO}(\theta, \eta, \alpha)$ is given by
\begin{equation}
    \mathcal{L}_\text{V-MPO}(\theta, \eta, \alpha) = \mathcal{L}_\pi(\theta) + \mathcal{L}_\eta(\eta) + \mathcal{L}_\alpha(\theta,\alpha).
\end{equation} Here the policy loss is the weighted maximum likelihood loss
\begin{equation}
    \mathcal{L}_\pi(\theta) = -\sum_{s,a\sim \tilde{\mathcal{D}}} \psi(s,a) \log \pi_\theta(a|s), \qquad 
    \psi(s,a) = \frac{\exp \big( \frac{A^\text{target}(s,a)}{\eta} \big)}{\sum_{s,a\sim \tilde{\mathcal{D}}} \exp \big( \frac{A^\text{target}(s,a)}{\eta} \big)}, \label{eq:policy_loss}
\end{equation} where the advantages $A^\text{target}(s,a)$ for the target network policy $\pi_{\theta_\text{target}}(a|s)$ are estimated according to the standard method described below. The tilde over the dataset, $\tilde{\mathcal{D}}$, indicates that we take samples corresponding to the top half advantages in the batch of data. The $\eta$, or ``temperature'', loss is
\begin{equation}
    \mathcal{L}_\eta(\eta) = \eta \epsilon_\eta + \eta \log \Bigg[ \frac{1}{|\tilde{\mathcal{D}}|}\sum_{s,a\sim \tilde{\mathcal{D}}} \exp\bigg( \frac{A^\text{target}(s,a)}{\eta} \bigg) \Bigg]. \label{eq:L_eta}
\end{equation} The KL constraint, which can be viewed as a form of trust-region loss, is given by
\begin{align}
    &\mathcal{L}_\alpha(\theta,\alpha)
    = \frac{1}{|\mathcal{D}|}\sum_{s \in \mathcal{D}} \Big[  \alpha \big( \epsilon_\alpha - \stopgradient\big[\big[ D_\text{KL}\big( \pi_{\theta_\text{target}}(a|s) \| \pi_{\vphantom{\theta_\text{target}}\theta}(a|s) \big) \big]\big] \Big) + \stopgradient[[\alpha]] D_\text{KL}\big( \pi_{\theta_\text{target}}(a|s) \| \pi_{\vphantom{\theta_\text{old}}\theta}(a|s) \big) \Big], \label{eq:L_alpha}
\end{align} where $\stopgradient[[\cdot]]$ indicates a stop gradient, i.e., that the enclosed term is assumed constant with respect to all variables. Note that here we use the full batch $\mathcal{D}$, not $\tilde{\mathcal{D}}$.

We used the Adam optimizer~\cite{Kingma2015} with default TensorFlow hyperparameters to optimize the total loss in Eq.~\ref{eq:total_loss}. In particular, the learning rate was fixed at $10^{-4}$ for all experiments.


\subsection{Policy evaluation}

In the present setting, policy evaluation means learning an approximate state-value function $V^\pi(s)$ given a policy $\pi(a|s)$, which we keep fixed for $T_\text{target}$ learning steps (i.e., batches of trajectories). We note that the value function corresponding to the target policy is instantiated in the ``online'' network receiving gradient updates; bootstrapping uses the online value function, as it is the best available estimate of the value function for the target policy. Thus in this section $\pi$ refers to $\pi_{\theta_\text{old}}$, while the value function update is performed on the current $\phi$, which may share parameters with the current $\theta$.

We fit a parametric value function $V^\pi_\phi(s)$ with parameters $\phi$ by minimizing the squared loss
\begin{align}
    \mathcal{L}_V(\phi) 
    &= \frac{1}{2|\mathcal{D}|}\sum_{s_t \sim \mathcal{D}} \big( V^\pi_\phi(s_t) - G^{(n)}_t \big)^2, \label{eq:value_loss}
\end{align} where $G^{(n)}_t$ is the standard $n$-step target for the value function at state $s_t$ at time $t$~\cite{Sutton1998}. This return uses the actual rewards in the trajectory and bootstraps from the value function for the rest: for each $\ell=t,\ldots,t+n-1$ in an unroll, $G^{(n)}_\ell = \sum_{k=\ell}^{t+n-1} \gamma^{k-\ell}r_k + \gamma^{t+n-\ell} V^\pi_\phi(s_{t+n})$. 
The advantages, which are the key quantity of interest for the policy improvement step in V-MPO, are then given by $A^\pi(s_t,a_t) = G^{(n)}_t - V^\pi_\phi(s_t)$ for each $s_t,a_t$ in the batch of trajectories.

\emph{PopArt normalization.} As we are interested in the multi-task setting where a single agent must learn a large number of tasks with differing reward scales, we used PopArt~\cite{Hasselt2016,Hessel2018} for the value function, even when training on a single task. Specifically, the value function outputs a separate value for each task in normalized space, which is converted to actual returns by a shift and scaling operation, the statistics of which are learned during training. We used a scale lower bound of $10^{-2}$, scale upper bound of $10^6$, and learning rate of $10^{-4}$ for the statistics. The lower bound guards against numerical issues when rewards are extremely sparse.

\emph{Importance-weighting for off-policy data.} It is possible to importance-weight the samples using V-trace to correct for off-policy data~\cite{Espeholt2018}, for example when data is taken from a replay buffer. For simplicity, however, no importance-weighting was used for the experiments presented in this work, which were mostly on-policy.


\subsection{Policy improvement in V-MPO}

In this section we show how, given the advantage function $A^{\pi_{\theta_\text{old}}}(s,a)$ for the state-action distribution $p_{\theta_\text{old}}(s,a)=\pi_{\theta_\text{old}}(a|s)p(s)$ induced by the old policy $\pi_{\theta_\text{old}}(a|s)$, we can estimate an improved policy $\pi_\theta(a|s)$. More formally, let $\mathcal{I}$ denote the binary event that the new policy is an improvement (in a sense to be defined below) over the previous policy: $\mathcal{I}=1$ if the policy is successfully improved and 0 otherwise. Then we would like to find the mode of the posterior distribution over parameters $\theta$ conditioned on this event, i.e., we seek the maximum a posteriori (MAP) estimate
\begin{equation}
    \theta^* = \operatorname{arg} \max_{\theta} \big[ \log p_{\theta}(\mathcal{I}=1) + \log p(\theta) \big], \label{eq:map_estimate}
\end{equation} where we have written $p(\mathcal{I}=1|\theta)$ as $p_\theta(\mathcal{I}=1)$ to emphasize the parametric nature of the dependence on $\theta$. We use the well-known identity $\log p(X) = \mathbb{E}_{\psi(Z)} \big[ \log \frac{p(X,Z)}{\psi(Z)} \big] + D_\text{KL}\big( \psi(Z) \| p(Z|X) \big)$ for any latent distribution $\psi(Z)$, where $D_\text{KL}(\psi(Z)\|p(Z|X))$ is the Kullback-Leibler divergence between $\psi(Z)$ and $p(Z|X)$ with respect to $Z$, and the first term is a lower bound because the KL divergence is always non-negative. Then considering $s,a$ as latent variables,
\begin{equation}
    \log p_\theta(\mathcal{I}=1) = \sum_{s,a} \psi(s,a) \log \frac{p_\theta(\mathcal{I}=1,s,a)}{\psi(s,a)} + D_\text{KL}\big( \psi(s,a) \| p_\theta(s,a|\mathcal{I}=1) \big). \label{eq:elbo}
\end{equation}

Policy improvement in V-MPO consists of the following two steps which have direct correspondences to the expectation maximization (EM) algorithm~\cite{Neal1998}: In the expectation (E) step, we choose the variational distribution $\psi(s,a)$ such that the lower bound on $\log p_\theta(\mathcal{I}=1)$ is as tight as possible, by minimizing the KL term. In the maximization (M) step we then find parameters $\theta$ that maximize the corresponding lower bound, together with the prior term in Eq.~\ref{eq:map_estimate}.


\subsubsection{E-step}

In the E-step, our goal is to choose the variational distribution $\psi(s,a)$ such that the lower bound on $\log p_\theta(\mathcal{I}=1)$ is as tight as possible, which is the case when the KL term in Eq.~\ref{eq:elbo} is zero. Given the old parameters $\theta_\text{old}$, this simply leads to $\psi(s,a) = p_{\theta_\text{old}}(s,a|\mathcal{I}=1)$, or
\begin{equation}
    \psi(s,a) = \frac{p_{\theta_\text{old}}(s,a)p_{\theta_\text{old}}(\mathcal{I}=1|s,a)}{p_{\theta_\text{old}}(\mathcal{I}=1)}, \qquad p_{\theta_\text{old}}(\mathcal{I}=1) = \sum_{s,a} p_{\theta_\text{old}}(s,a)p_{\theta_\text{old}}(\mathcal{I}=1|s,a). \label{eq:psi_general}
\end{equation} Intuitively, this solution weights the probability of each state-action pair with its relative improvement probability $p_{\theta_\text{old}}(\mathcal{I}=1|s,a)$. We now choose a distribution $p_{\theta_\text{old}}(\mathcal{I}=1|s,a)$ that leads to our desired outcome. As we prefer actions that lead to a higher advantage in each state, we suppose that this probability is given by
\begin{equation}
    p_{\theta_\text{old}}(\mathcal{I}=1|s,a) \propto \exp \bigg( \frac{A^{\pi_{\theta_\text{old}}}(s,a)}{\eta} \bigg) \label{eq:improvement_criterion}
\end{equation} for some temperature $\eta > 0$, from which we obtain the equation on the right in Eq.~\ref{eq:policy_loss}. This probability depends on the \emph{old} parameters $\theta_\text{old}$ and not on the new parameters $\theta$. Meanwhile, the value of $\eta$ allows us to control the diversity of actions that contribute to the weighting, but at the moment is arbitrary. It turns out, however, that we can tune $\eta$ as part of the optimization, which is desirable since the optimal value of $\eta$ changes across iterations. The convex loss that achieves this, Eq.~\ref{eq:L_eta}, is derived in Appendix~\ref{appendix:temperature_loss} by minimizing the KL term in Eq.~\ref{eq:elbo} subject to a hard constraint on $\psi(s,a)$.

\emph{Top-$k$ advantages.} We found that learning improves substantially if we take only the samples corresponding to the highest 50\% of advantages in each batch for the E-step, corresponding to the use of $\tilde{\mathcal{D}}$ rather than $\mathcal{D}$ in Eqs.~\ref{eq:policy_loss},~\ref{eq:L_eta}. Importantly, these must be consistent between the maximum likelihood weights in Eq.~\ref{eq:policy_loss} and the temperature loss in Eq.~\ref{eq:L_eta}, since, mathematically, this is justified by choosing the corresponding policy improvement probability in Eq.~\ref{eq:improvement_criterion} to only use the top half of the advantages. This is similar to the technique used in Covariance Matrix Adaptation - Evolutionary Strategy (CMA-ES)~\cite{Hansen1997,Abdolmaleki2017}, and is a special case of the more general feature that any rank-preserving transformation is allowed under this formalism.

\emph{Importance weighting for off-policy corrections.} As for the value function, importance weights can be used in the policy improvement step to correct for off-policy data. While not used for the experiments presented in this work, details for how to carry out this correction are given in Appendix~\ref{appendix:importance_weighting}.


\subsubsection{M-step: Constrained supervised learning of the parametric policy}

In the E-step we found the nonparametric variational state-action distribution $\psi(s,a)$, Eq.~\ref{eq:psi_general}, that gives the tightest lower bound to $p_\theta(\mathcal{I}=1)$ in Eq.~\ref{eq:elbo}. In the M-step we maximize this lower bound together with the prior term $\log p(\theta)$ with respect to the parameters $\theta$, which effectively leads to a constrained weighted maximum likelihood problem. Thus the introduction of the nonparametric distribution in Eq.~\ref{eq:psi_general} separates the RL procedure from the neural network fitting.

We would like to find new parameters $\theta$ that minimize
\begin{equation}
    \mathcal{L}(\theta) 
    = -\sum_{s,a} \psi(s,a) \log \frac{p_\theta(\mathcal{I}=1,s,a)}{\psi(s,a)} - \log p(\theta). \label{eq:original_m_step_loss}
\end{equation} Note, however, that so far we have worked with the joint state-action distribution $\psi(s,a)$ while we are in fact optimizing for the policy, which is the conditional distribution $\pi_\theta(a|s)$. Writing $p_\theta(s,a) = \pi_\theta(a|s)p(s)$ since only the policy is parametrized by $\theta$ and dropping terms that are not parametrized by $\theta$, the first term of Eq.~\ref{eq:original_m_step_loss} is seen to be the weighted maximum likelihood policy loss
\begin{equation}
   \mathcal{L}_\pi(\theta) = -\sum_{s,a} \psi(s,a) \log \pi_\theta(a|s).
\end{equation} In the sample-based computation of this loss, we assume that any state-action pairs not in the batch of trajectories have zero weight, leading to the normalization in Eq.~\ref{eq:policy_loss}.

As in the original MPO algorithm, a useful prior is to keep the new policy $\pi_\theta(a|s)$ close to the old policy $\pi_{\theta_\text{old}}(a|s)$: $\log p(\theta) \approx -\alpha \mathbb{E}_{s\sim p(s)} \big[ \dkl\big( \pi_{\theta_\text{old}}(a|s) \| \pi_{\vphantom{\theta_\text{old}}\theta}(a|s) \big) \big]$. While intuitive, we motivate this more formally in Appendix~\ref{appendix:kl-constraint}. It is again more convenient to specify a bound on the KL divergence instead of tuning $\alpha$ directly, so we solve the constrained optimization problem
\begin{equation}
    \theta^* = \argmin{\theta} - \sum_{s,a} \psi(s,a) \log \pi_{\theta}(a|s) \quad \text{s.t. } \operatornamewithlimits{\mathbb{E}}_{s\sim p(s)} \Big[ D_\text{KL}\big( \pi_{\theta_\text{old}}(a|s) \| \pi_\theta(a|s) \big) \Big] < \epsilon_\alpha. \label{eq:m_step_constraint}
\end{equation} Intuitively, the constraint in the E-step expressed by Eq.~\ref{eq:e_step_constraint} in Appendix~\ref{appendix:temperature_loss} for tuning the temperature only constrains the nonparametric distribution; it is the constraint in Eq.~\ref{eq:m_step_constraint} that directly limits the change in the parametric policy, in particular for states and actions that were not in the batch of samples and which rely on the generalization capabilities of the neural network function approximator.

To make the constrained optimization problem amenable to gradient descent, we use Lagrangian relaxation to write the unconstrained objective as
\begin{equation}
    \mathcal{J}(\theta, \alpha)
    = \mathcal{L}_\pi(\theta) + \alpha \bigg( \epsilon_\alpha - \operatornamewithlimits{\mathbb{E}}_{s\sim p(s)} \Big[ D_\text{KL} \big( \pi_{\theta_\text{old}}(a|s) \| \pi_{\vphantom{\theta_\text{old}}\theta}(a|s) \big) \Big] \bigg), \label{eq:m_step_objective}
\end{equation} which we can optimize by following a coordinate-descent strategy, alternating between the optimization over $\theta$ and $\alpha$. Thus, in addition to the policy loss we arrive at the constraint loss
\begin{equation}
    \mathcal{L}_\alpha(\theta,\alpha) = \alpha \bigg( \epsilon_\alpha - \operatornamewithlimits{\mathbb{E}}_{s\sim p(s)} \Big[ \stopgradient\big[\big[ D_\text{KL}\big( \pi_{\theta_\text{old}} \| \pi_{\vphantom{\theta_\text{old}}\theta} \big) \big]\big] \Big] \bigg) + \stopgradient[[\alpha]] \operatornamewithlimits{\mathbb{E}}_{s\sim p(s)} \Big[ D_\text{KL}\big( \pi_{\theta_\text{old}} \| \pi_{\vphantom{\theta_\text{old}}\theta} \big) \Big]. \label{eq:m_step_kl_constraint}
\end{equation} Replacing the sum over states with samples gives Eq.~\ref{eq:L_alpha}. Since $\eta$ and $\alpha$ are Lagrange multipliers that must be positive, after each gradient update we project the resulting $\eta$ and $\alpha$ to a small positive value which we choose to be $\eta_\text{min}=\alpha_\text{min}=10^{-8}$ throughout the results presented below.

For continuous action spaces parametrized by Gaussian distributions, we use decoupled KL constraints for the M-step in Eq.~\ref{eq:m_step_kl_constraint} as in \citeauthor{Abdolmaleki2018}~(\citeyear{Abdolmaleki2018}); the precise form is given in Appendix~\ref{appendix:continuous_control}.

\section{Experiments}

Details on the network architecture and hyperparameters used for each task are given in Appendix~\ref{appendix:network_architecture}.

\subsection{Discrete actions: DMLab, Atari}

\begin{figure}[h]
\begin{center}
\subcaptionbox{Multi-task DMLab-30.}{\includegraphics[trim={1.5cm 0 1.2cm 0},clip,width=0.35\linewidth]{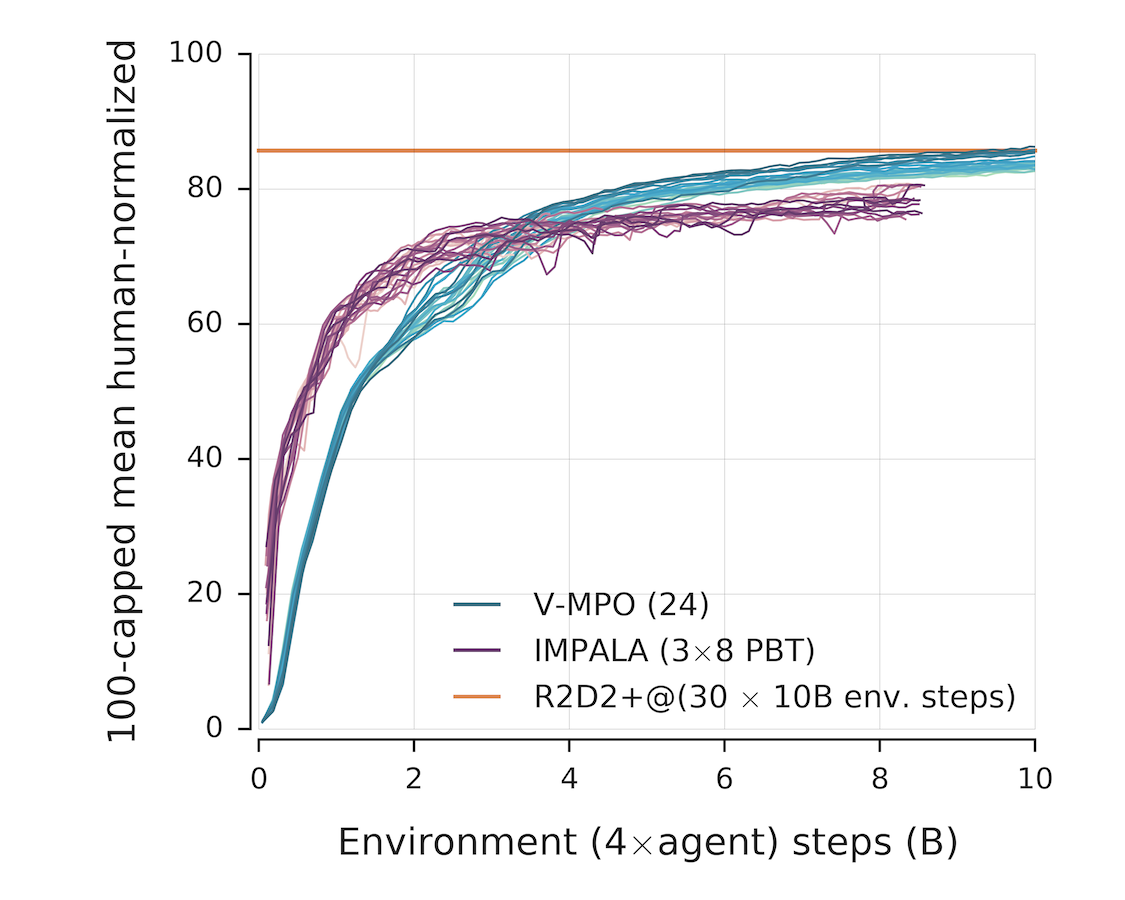}}
\hspace*{0.03\columnwidth}
\subcaptionbox{Multi-task Atari-57.}{\includegraphics[trim={1.5cm 0 1.2cm 0},clip,width=0.35\columnwidth]{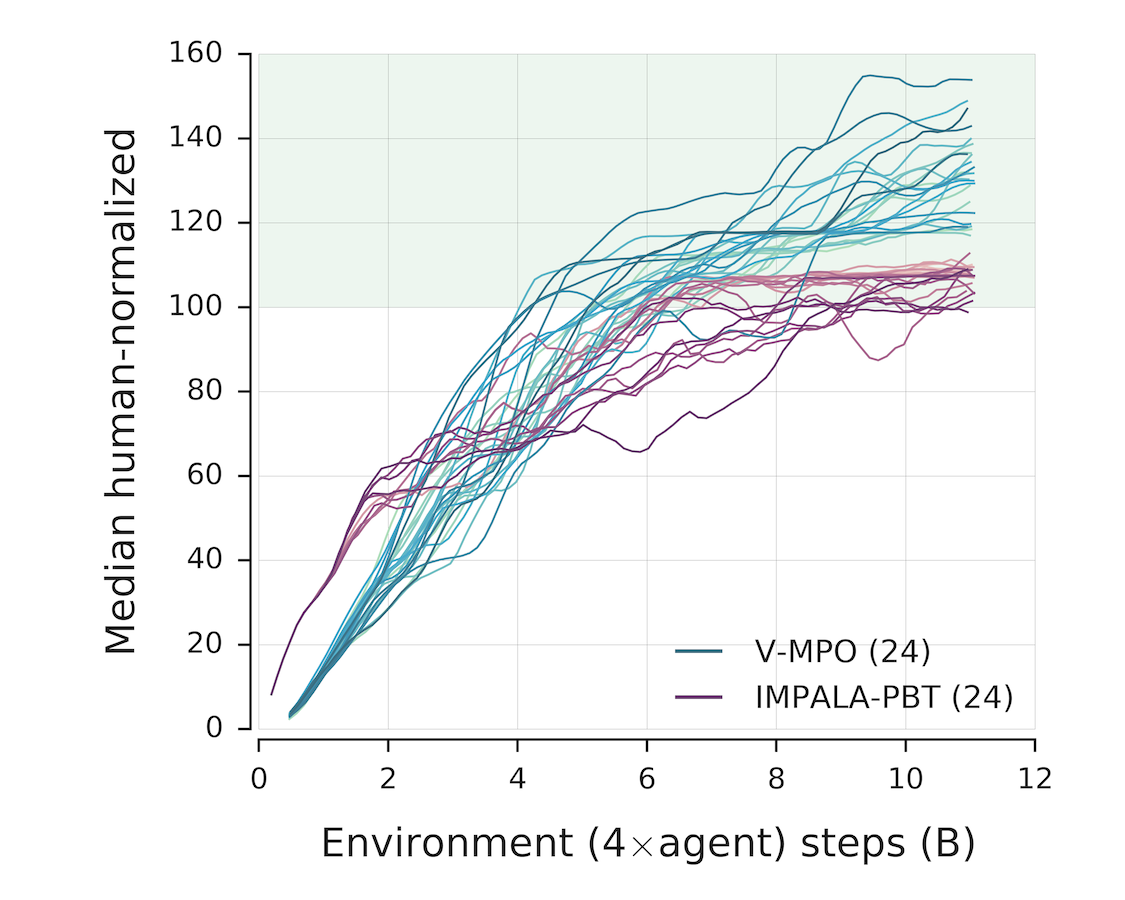}}
\caption{(a) Multi-task DMLab-30. IMPALA results show 3 runs of 8 agents each; within a run hyperparameters were evolved via PBT. For V-MPO each line represents a set of hyperparameters that are fixed throughout training. The final result of R2D2+ trained for 10B environment steps on individual levels~\cite{Kapturowski2019} is also shown for comparison (orange line). (b) Multi-task Atari-57. In the IMPALA experiment, hyperparameters were evolved with PBT. For V-MPO each of the 24 lines represents a set of hyperparameters that were fixed throughout training, and all runs achieved a higher score than the best IMPALA run. Data for IMPALA (``Pixel-PopArt-IMPALA'' for DMLab-30 and ``PopArt-IMPALA'' for Atari-57) was obtained from the authors of \citeauthor{Hessel2018}~(\citeyear{Hessel2018}). Each environment frame corresponds to 4 agent steps due to the action repeat.}
\label{fig:dmlab-30}
\end{center}
\end{figure}

\begin{figure}[h]
\begin{center}
\includegraphics[trim={2.1cm 0 1.2cm 0},clip,width=0.245\linewidth]{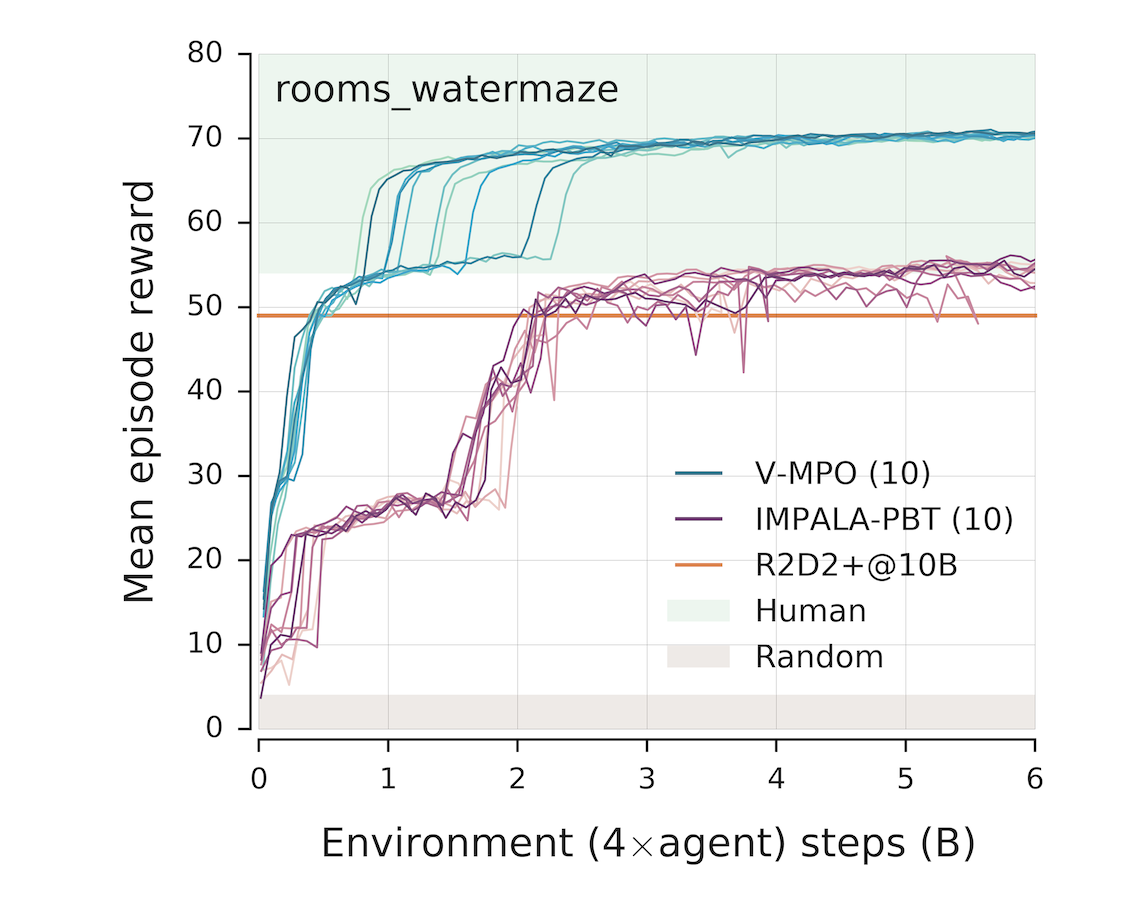}
\includegraphics[trim={2.1cm 0 1.2cm 0},clip,width=0.245\linewidth]{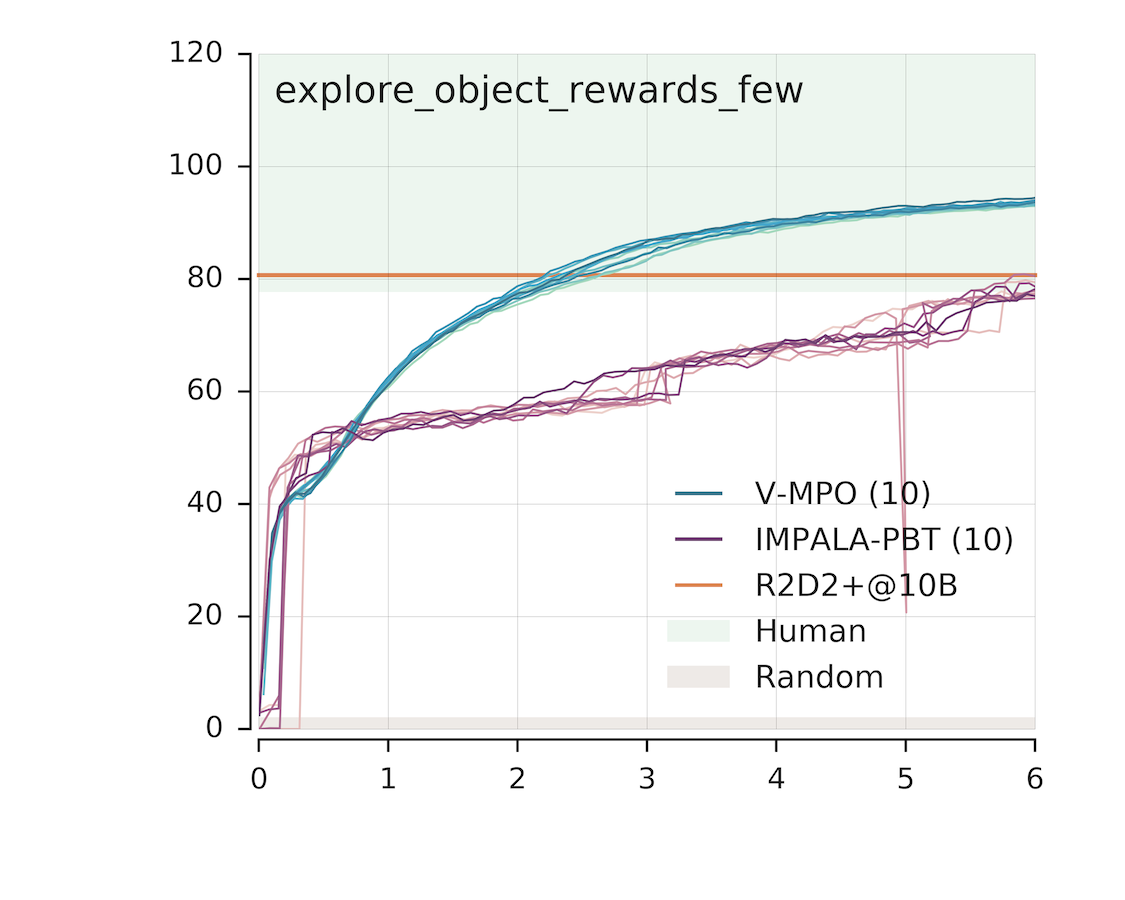}
\includegraphics[trim={2.1cm 0 1.2cm 0},clip,width=0.245\linewidth]{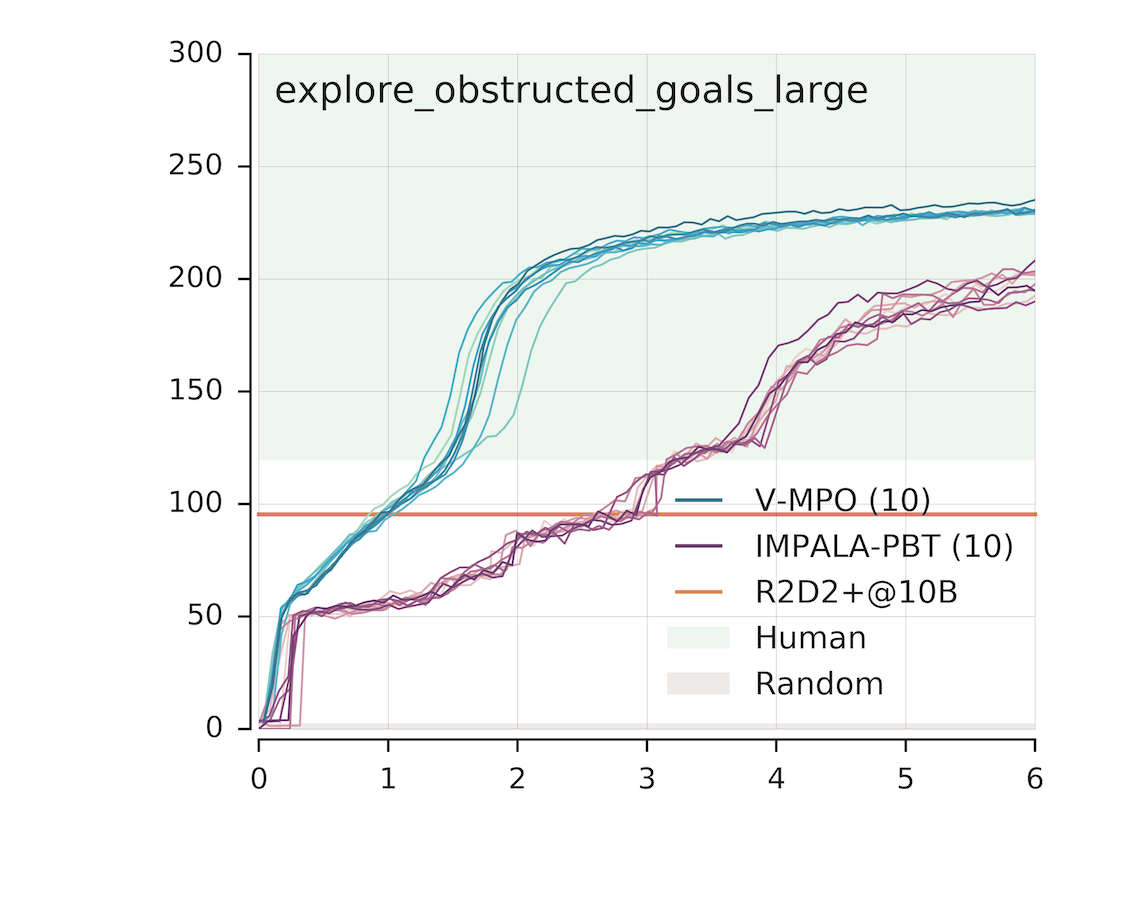}
\includegraphics[trim={2.1cm 0 1.2cm 0},clip,width=0.245\linewidth]{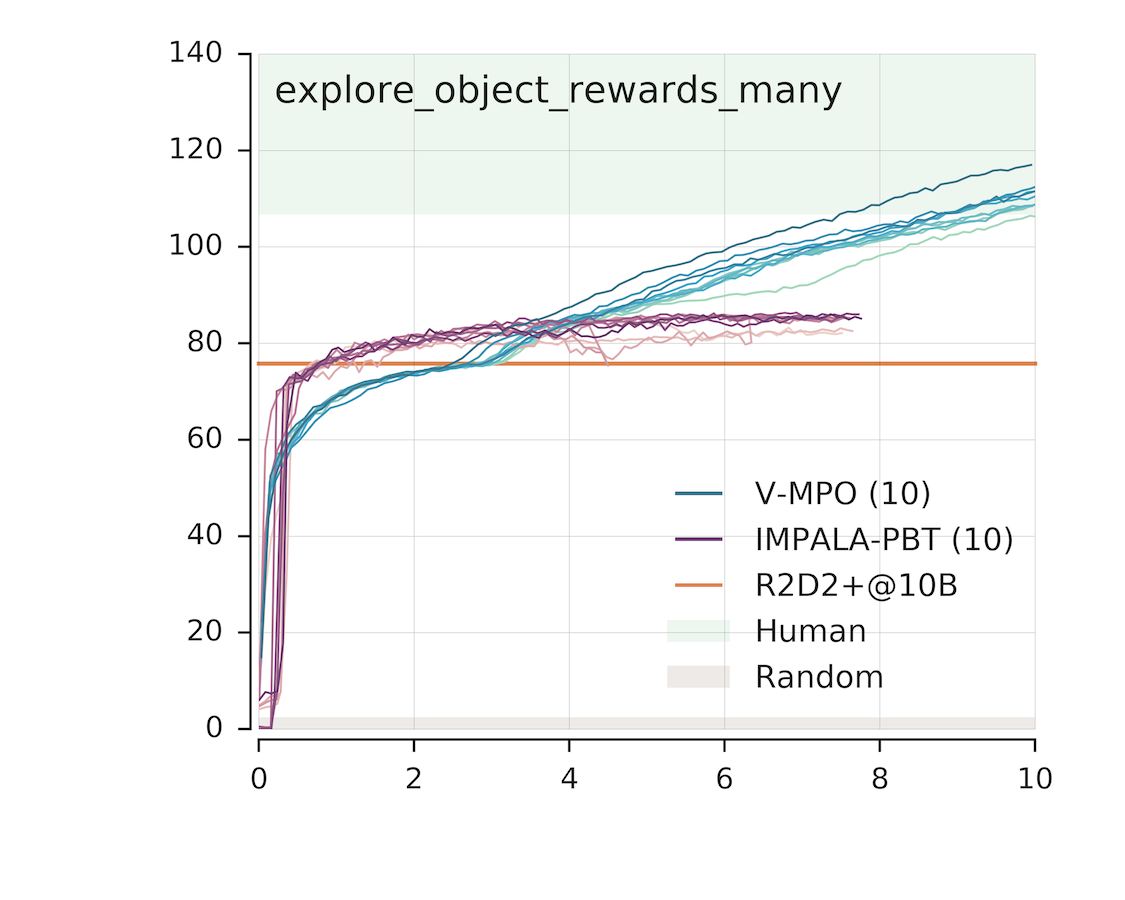}
\caption{Example levels from DMLab-30, compared to IMPALA and more recent results from R2D2+, the larger, DMLab-specific version of R2D2~\cite{Kapturowski2019}. The IMPALA results include hyperparameter evolution with PBT.}
\label{fig:dmlab_example_levels}
\end{center}
\end{figure}

\begin{figure}[h]
\begin{center}
\includegraphics[trim={1.4cm 0 1.0cm 0},clip,width=0.245\linewidth]{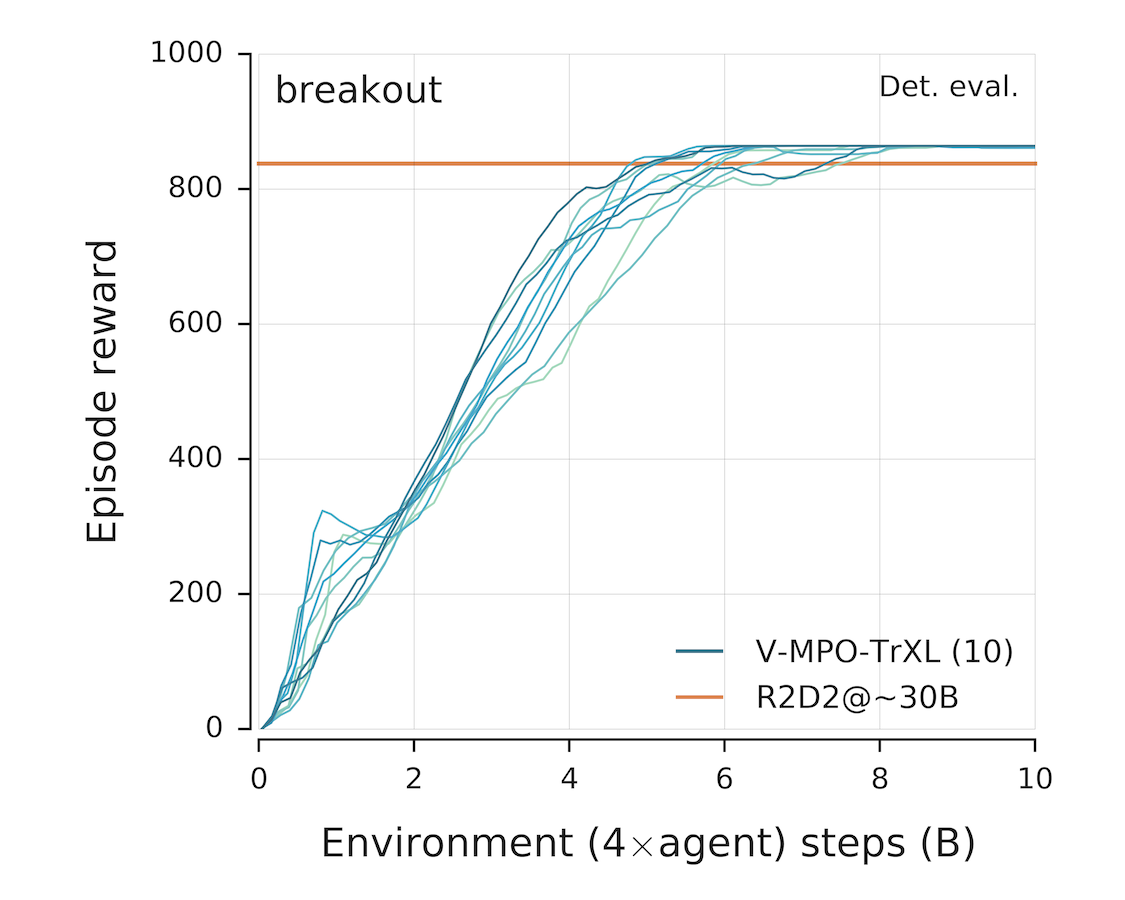}
\includegraphics[trim={1.4cm 0 1.0cm 0},clip,width=0.245\linewidth]{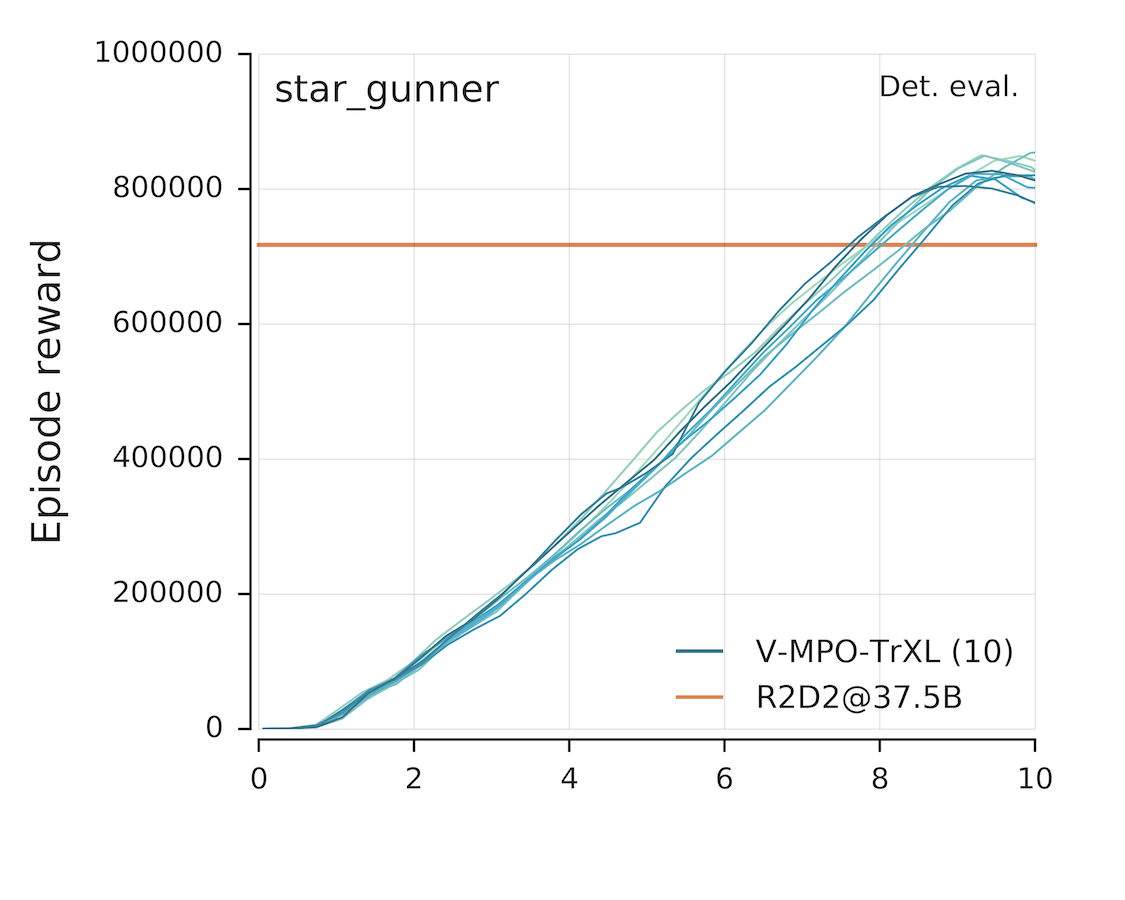}
\includegraphics[trim={1.4cm 0 1.0cm 0},clip,width=0.245\linewidth]{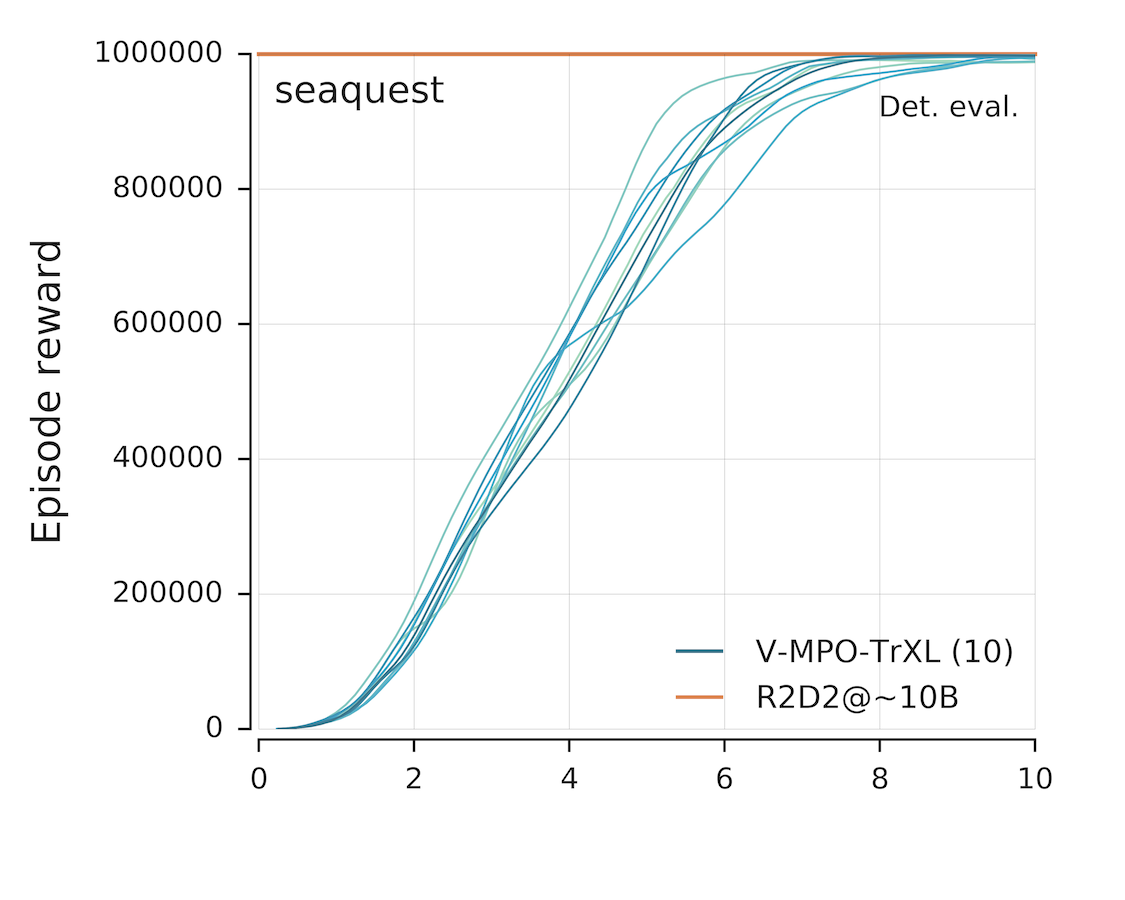}
\includegraphics[trim={1.4cm 0 1.0cm 0},clip,width=0.245\linewidth]{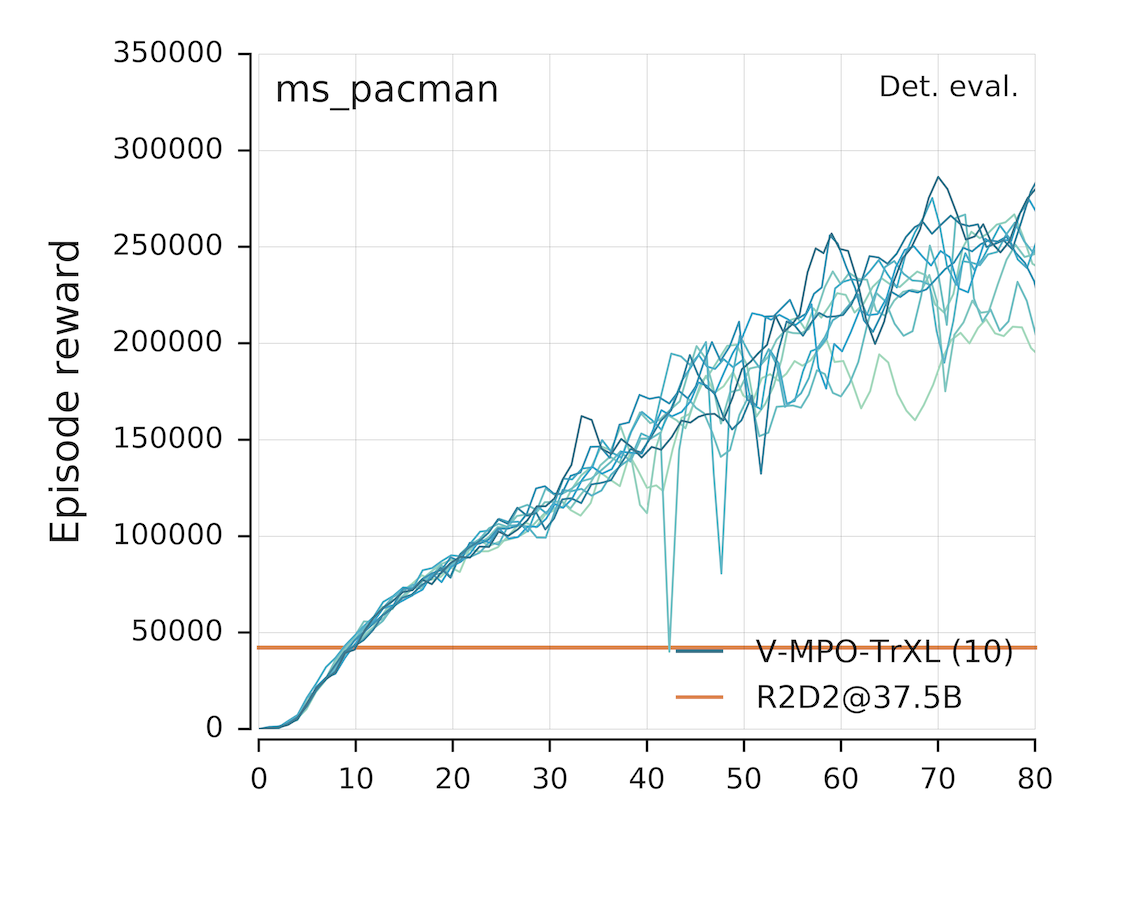}
\caption{Example levels from Atari. In Breakout, V-MPO achieves the maximum score of 864 in every episode. No reward clipping was applied, and the maximum length of an episode was 30 minutes (108,000 frames). Supplementary video for Ms. Pacman: \url{https://bit.ly/2lWQBy5}}
\label{fig:atari_example_levels}
\end{center}
\end{figure}

\begin{figure}[t]
\begin{center}
\subcaptionbox{}{\includegraphics[trim={0 0 0cm 0},clip,width=0.245\linewidth]{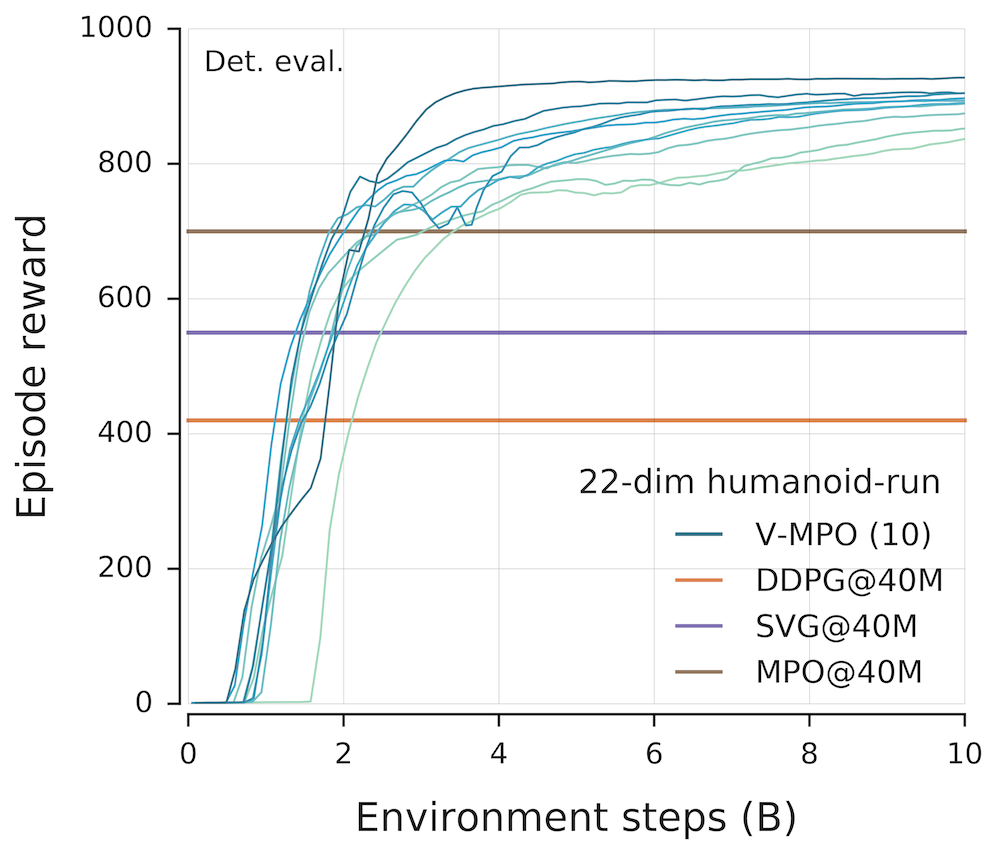}}
\subcaptionbox{}{\includegraphics[trim={0cm 0 0cm 0},clip,width=0.245\linewidth]{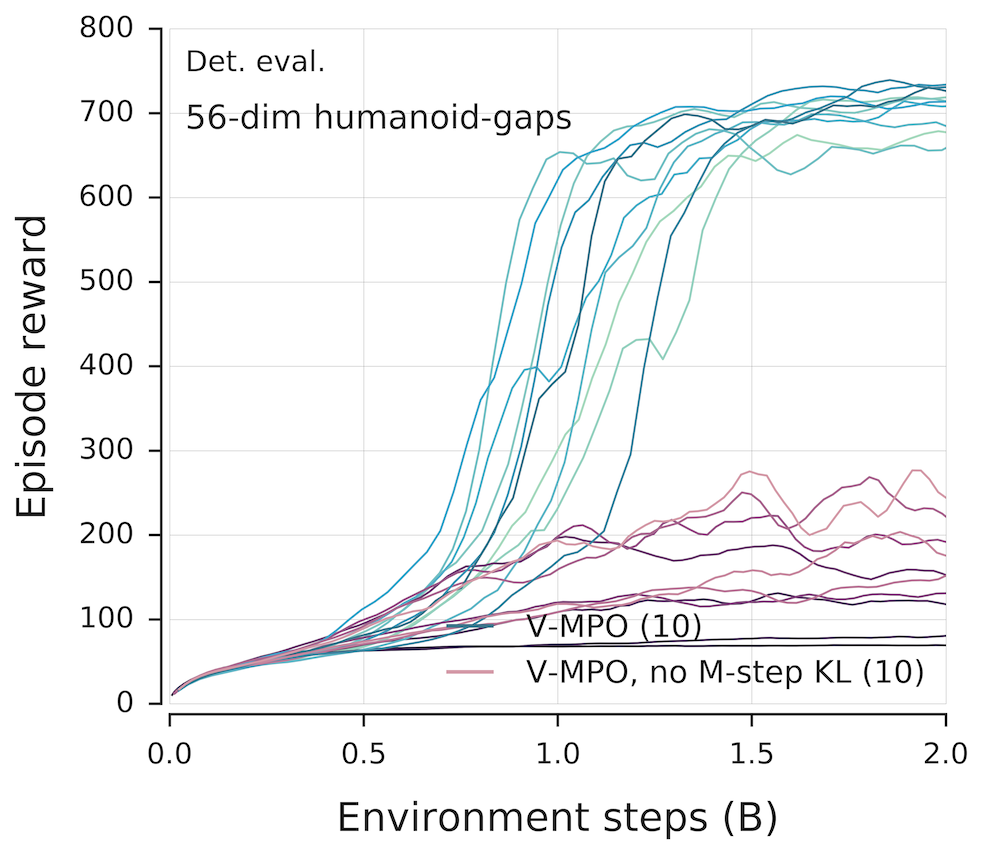}}
\subcaptionbox{}{\includegraphics[trim={0cm 0 0cm 0},clip,width=0.245\linewidth]{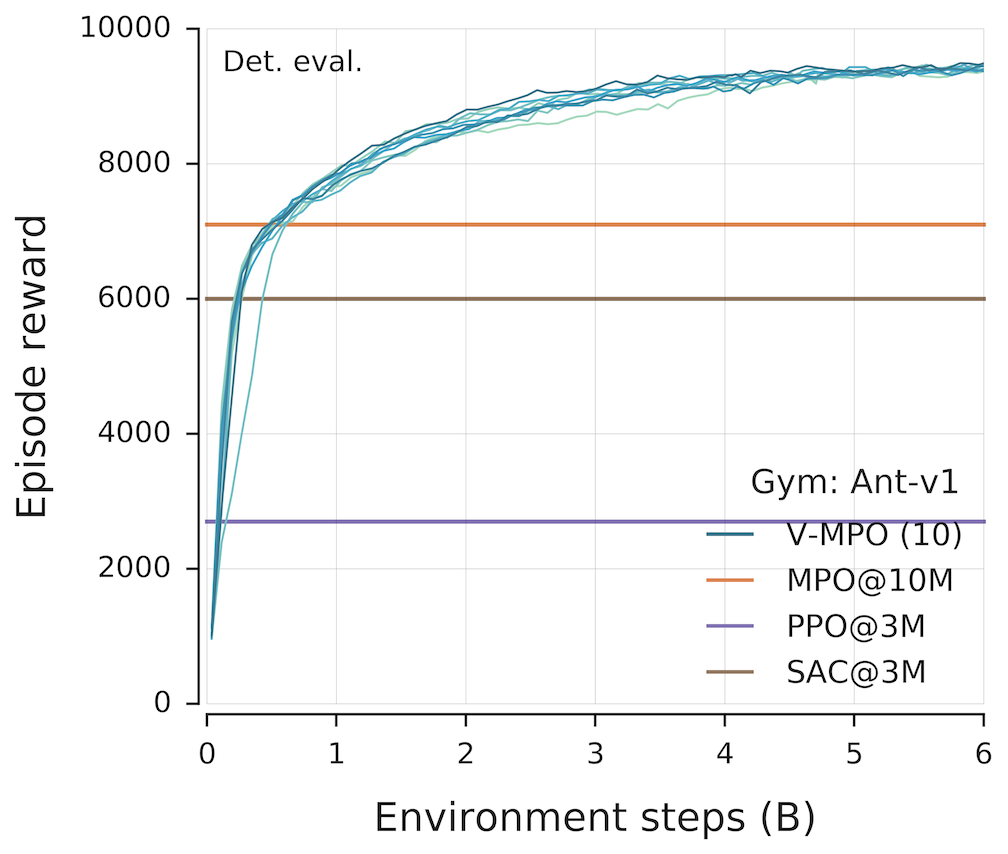}}
\subcaptionbox{}{\includegraphics[trim={0cm 0 0cm 0},clip,width=0.245\linewidth]{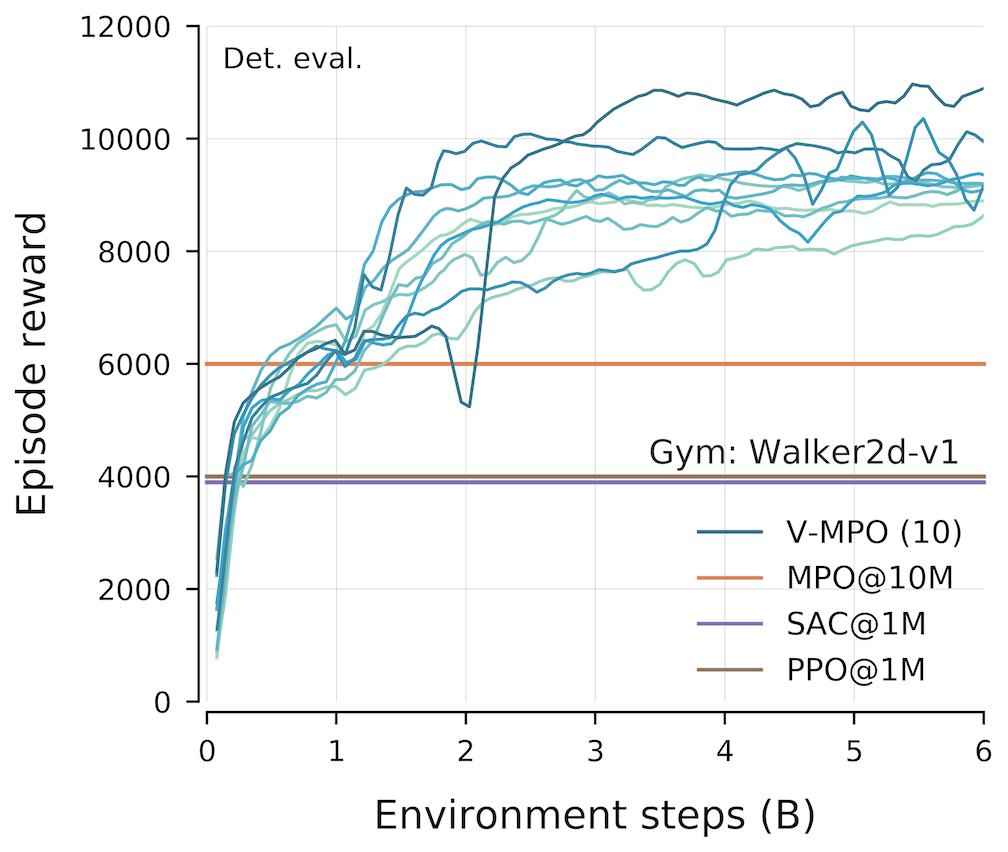}}

\caption{(a) Humanoid ``run'' from full state~\cite{Tassa2018} and (b) humanoid ``gaps'' from pixel observations~\cite{Merel2019}. Purple curves are the same runs but without parametric KL constraints. Det. eval.: deterministic evaluation. Supplementary video for humanoid gaps: \url{https://bit.ly/2L9KZdS}. (c)-(d) Example OpenAI Gym tasks.}
\label{fig:humanoid_gaps}
\end{center}
\end{figure}

\emph{DMLab.} DMLab-30~\cite{Beattie2016} is a collection of visually rich, partially observable 3D environments played from the first-person point of view. Like IMPALA, for DMLab we used pixel control as an auxiliary loss for representation learning~\cite{Jaderberg2016,Hessel2018}. However, we did not employ the optimistic asymmetric reward scaling used by previous IMPALA experiments to aid exploration on a subset of the DMLab levels, by weighting positive rewards more than negative rewards~\cite{Espeholt2018, Hessel2018,Kapturowski2019}. Unlike in \citeauthor{Hessel2018}~(\citeyear{Hessel2018}) we also did not use population-based training (PBT)~\cite{Jaderberg2017}. Additional details for the settings used in DMLab can be found in Table~\ref{table:dmlab_settings} of the Appendix.

Fig.~\ref{fig:dmlab-30}a shows the results for multi-task DMLab-30, comparing the V-MPO learning curves to data obtained from \citeauthor{Hessel2018}~(\citeyear{Hessel2018}) for the PopArt IMPALA agent with pixel control. We note that the result for V-MPO at 10B environment frames across all levels matches the result for the Recurrent Replay Distributed DQN (R2D2) agent~\cite{Kapturowski2019} trained on \emph{individual levels} for 10B environment steps per level. Fig.~\ref{fig:dmlab_example_levels} shows example individual levels in DMLab where V-MPO achieves scores that are substantially higher than has previously been reported, for both R2D2 and IMPALA. The pixel-control IMPALA agents shown here were carefully tuned for DMLab and are similar to the ``experts'' used in \citeauthor{Schmitt2018}~(\citeyear{Schmitt2018}); in all cases these results match or exceed previously published results for IMPALA~\cite{Espeholt2018,Kapturowski2019}.

\emph{Atari.} The Atari Learning Environment (ALE)~\cite{Bellemare2012} is a collection of 57 Atari 2600 games that has served as an important benchmark for recent deep RL methods. We used the standard preprocessing scheme and a maximum episode length of 30 minutes (108,000 frames), see Table~\ref{table:atari_settings} in the Appendix. For the multi-task setting we followed \citeauthor{Hessel2018}~(\citeyear{Hessel2018}) in setting the discount to zero on loss of life; for the example single tasks we did not employ this trick, since it can prevent the agent from achieving the highest score possible by sacrificing lives. Similarly, while in the multi-task setting we followed previous work in clipping the maximum reward to 1.0, no such clipping was applied in the single-task setting in order to preserve the original reward structure. Additional details for the settings used in Atari can be found in Table~\ref{table:atari_settings} in the Appendix.

Fig.~\ref{fig:dmlab-30}b shows the results for multi-task Atari-57, demonstrating that it is possible for a \emph{single} agent to achieve ``superhuman`` median performance on Atari-57 in approximately 4 billion ($\sim$70 million per level) environment frames.

We also compare the performance of V-MPO on a few individual Atari levels to R2D2~\cite{Kapturowski2019}, which previously achieved some of the highest scores reported for Atari. Again, V-MPO can match or exceed previously reported scores while requiring fewer interactions with the environment. In Ms. Pacman, the final performance approaches 300,000 with a 30-minute timeout (and the maximum 1M without), effectively solving the game. Inspired by the argument in \citeauthor{Kapturowski2019}~(\citeyear{Kapturowski2019}) that in a fully observable environment LSTMs enable the agent to utilize more useful representations than is available in the immediate observation, for the single-task setting we used a Transformer-XL (TrXL)~\cite{Dai2019} to replace the LSTM core. Unlike previous work for single Atari levels, we did not employ any reward clipping~\cite{Mnih2015,Espeholt2018} or nonlinear value function rescaling~\cite{Kapturowski2019}.

\subsection{Continuous control}

To demonstrate V-MPO's effectiveness in high-dimensional, continuous action spaces, here we present examples of learning to control both a simulated humanoid with 22 degrees of freedom from full state observations and one with 56 degrees of freedom from pixel observations~\cite{Tassa2018,Merel2019}. As shown in Fig.~\ref{fig:humanoid_gaps}a, for the 22-dimensional humanoid V-MPO reliably achieves higher asymptotic returns than has previously been reported, including for Deep Deterministic Policy Gradients (DDPG)~\cite{Lillicrap2015}, Stochastic Value Gradients (SVG)~\cite{Heess2015}, and MPO. These algorithms are far more sample-efficient but reach a lower final performance.

In the ``gaps'' task the 56-dimensional humanoid must run forward to match a target velocity of 4~m/s and jump over the gaps between platforms by learning to actuate joints with position-control~\cite{Merel2019}. Previously, only an agent operating in the space of pre-learned motor primitives was able to solve the task from pixel observations~\cite{Merel2018,Merel2019}; here we show that V-MPO can learn a challenging visuomotor task \emph{from scratch} (Fig.~\ref{fig:humanoid_gaps}b). For this task we also demonstrate the importance of the parametric KL constraint, without which the agent learns poorly.

In Figs.~\ref{fig:humanoid_gaps}c-d we also show that V-MPO achieves the highest asymptotic performance reported for two OpenAI Gym tasks~\cite{Brockman2016}. Again, MPO and Stochastic Actor-Critic~\cite{Haarnoja2018} are far more sample-efficient but reach a lower final performance.

\section{Conclusion}

In this work we have introduced a scalable on-policy deep reinforcement learning algorithm, V-MPO, that is applicable to both discrete and continuous control domains. For the results presented in this work neither importance weighting nor entropy regularization was used; moreover, since the size of neural network parameter updates is limited by KL constraints, we were also able to use the same learning rate for all experiments. This suggests that a scalable, performant RL algorithm may not require some of the tricks that have been developed over the past several years. Interestingly, both the original MPO algorithm for replay-based off-policy learning~\cite{Abdolmaleki2018a,Abdolmaleki2018} and V-MPO for on-policy learning are derived from similar principles, providing evidence for the benefits of this approach as an alternative to popular policy gradient-based methods.


\subsubsection*{Acknowledgments}
We thank Lorenzo Blanco, Trevor Cai, Greg Wayne, Chloe Hillier, and Vicky Langston for their assistance and support.



\begin{thebibliography}{45}
\providecommand{\natexlab}[1]{#1}
\providecommand{\url}[1]{\texttt{#1}}
\expandafter\ifx\csname urlstyle\endcsname\relax
  \providecommand{\doi}[1]{doi: #1}\else
  \providecommand{\doi}{doi: \begingroup \urlstyle{rm}\Url}\fi

\bibitem[Abdolmaleki et~al.(2017)Abdolmaleki, Price, Lau, Reis, and
  Neumann]{Abdolmaleki2017}
Abbas Abdolmaleki, Bob Price, Nuno Lau, Luis~P Reis, and Gerhard Neumann.
\newblock {Deriving and Improving CMA-ES with Information Geometric Trust
  Regions}.
\newblock \emph{Proceedings of the Genetic and Evolutionary Computation
  Conference}, 2017.

\bibitem[Abdolmaleki et~al.(2018{\natexlab{a}})Abdolmaleki, Springenberg,
  Degrave, Bohez, Tassa, Belov, Heess, and Riedmiller]{Abdolmaleki2018a}
Abbas Abdolmaleki, Jost~Tobias Springenberg, Jonas Degrave, Steven Bohez, Yuval
  Tassa, Dan Belov, Nicolas Heess, and Martin Riedmiller.
\newblock {Relative Entropy Regularized Policy Iteration}.
\newblock \emph{arXiv preprint}, 2018{\natexlab{a}}.
\newblock URL \url{https://arxiv.org/pdf/1812.02256.pdf}.

\bibitem[Abdolmaleki et~al.(2018{\natexlab{b}})Abdolmaleki, Springenberg,
  Tassa, Munos, Heess, and Riedmiller]{Abdolmaleki2018}
Abbas Abdolmaleki, Jost~Tobias Springenberg, Yuval Tassa, Remi Munos, Nicolas
  Heess, and Martin Riedmiller.
\newblock {Maximum a Posteriori Policy Optimisation}.
\newblock \emph{Int. Conf. Learn. Represent.}, 2018{\natexlab{b}}.
\newblock URL \url{https://arxiv.org/pdf/1806.06920.pdf}.

\bibitem[{Anonymous Authors}(2019)]{Schmitt2019}
{Anonymous Authors}.
\newblock {Off-Policy Actor-Critic with Shared Experience Replay}.
\newblock \emph{Under review, Int. Conf. Learn. Represent.}, 2019.

\bibitem[Beattie et~al.(2016)Beattie, Leibo, Teplyashin, Ward, Wainwright,
  K{\"u}ttler, Lefrancq, Green, Vald{\'e}s, Sadik, et~al.]{Beattie2016}
Charles Beattie, Joel~Z Leibo, Denis Teplyashin, Tom Ward, Marcus Wainwright,
  Heinrich K{\"u}ttler, Andrew Lefrancq, Simon Green, V{\'\i}ctor Vald{\'e}s,
  Amir Sadik, et~al.
\newblock {Deepmind Lab}.
\newblock \emph{arXiv preprint arXiv:1612.03801}, 2016.

\bibitem[Bellemare et~al.(2012)Bellemare, Naddaf, Veness, and
  Bowling]{Bellemare2012}
Marc~G. Bellemare, Yavar Naddaf, Joel Veness, and Michael Bowling.
\newblock {The Arcade Learning Environment: An Evaluation Platform for General
  Agents}.
\newblock \emph{Journal of Artificial Intelligence Research}, 47, 2012.

\bibitem[Brockman et~al.(2016)Brockman, Cheung, Pettersson, Schneider,
  Schulman, Tang, and Zaremba]{Brockman2016}
Greg Brockman, Vicki Cheung, Ludwig Pettersson, Jonas Schneider, John Schulman,
  Jie Tang, and Wojciech Zaremba.
\newblock {OpenAI Gym}.
\newblock \emph{arXiv preprint}, 2016.
\newblock URL \url{http://arxiv.org/abs/1606.01540}.

\bibitem[Buchlovsky et~al.(2019)Buchlovsky, Budden, Grewe, Jones, Aslanides,
  Besse, Brock, Clark, Colmenarejo, Pope, Viola, and Belov]{Buchlovsky2019}
Peter Buchlovsky, David Budden, Dominik Grewe, Chris Jones, John Aslanides,
  Frederic Besse, Andy Brock, Aidan Clark, Sergio~Gomez Colmenarejo, Aedan
  Pope, Fabio Viola, and Dan Belov.
\newblock {TF-Replicator: Distributed Machine Learning for Researchers}.
\newblock \emph{arXiv preprint}, 2019.
\newblock URL \url{http://arxiv.org/abs/1902.00465}.

\bibitem[Dai et~al.(2019)Dai, Yang, Yang, Carbonell, Le, and
  Salakhutdinov]{Dai2019}
Zihang Dai, Zhilin Yang, Yiming Yang, Jaime~G. Carbonell, Quoc~V. Le, and
  Ruslan Salakhutdinov.
\newblock {Transformer-XL: Attentive Language Models Beyond a Fixed-Length
  Context}.
\newblock \emph{arXiv preprint}, 2019.
\newblock URL \url{http://arxiv.org/abs/1901.02860}.

\bibitem[{DeepMind}(2019)]{DeepMind2019}
{DeepMind}.
\newblock {AlphaStar: Mastering the Real-Time Strategy Game StarCraft II},
  2019.
\newblock URL
  \url{https://deepmind.com/blog/alphastar-mastering-real-time-strategy-game-starcraft-ii/}.

\bibitem[Duan et~al.(2016)Duan, Chen, Houthooft, Schulman, and
  Abbeel]{Duan2018}
Yan Duan, Xi~Chen, Rein Houthooft, John Schulman, and Pieter Abbeel.
\newblock {Benchmarking Deep Reinforcement Learning for Continuous Control}.
\newblock \emph{arXiv preprint}, 2016.
\newblock URL \url{http://arxiv.org/abs/1604.06778}.

\bibitem[Espeholt et~al.(2018)Espeholt, Soyer, Munos, Simonyan, Mnih, Ward,
  Doron, Firoiu, Harley, Dunning, Legg, and Kavukcuoglu]{Espeholt2018}
Lasse Espeholt, Hubert Soyer, Remi Munos, Karen Simonyan, Volodymir Mnih, Tom
  Ward, Yotam Doron, Vlad Firoiu, Tim Harley, Iain Dunning, Shane Legg, and
  Koray Kavukcuoglu.
\newblock {IMPALA: Scalable Distributed Deep-RL with Importance Weighted
  Actor-Learner Architectures}.
\newblock \emph{arXiv preprint}, 2018.
\newblock URL \url{http://arxiv.org/abs/1802.01561}.

\bibitem[{Google}(2018)]{Google2018}
{Google}.
\newblock {Cloud TPU}, 2018.
\newblock URL \url{https://cloud.google.com/tpu/}.

\bibitem[Haarnoja et~al.(2018)Haarnoja, Zhou, Abbeel, and Levine]{Haarnoja2018}
Tuomas Haarnoja, Aurick Zhou, Pieter Abbeel, and Sergey Levine.
\newblock {Soft Actor-Critic: Off-Policy Maximum Entropy Deep Reinforcement
  Learning with a Stochastic Actor}.
\newblock \emph{arXiv preprint}, 2018.
\newblock URL \url{http://arxiv.org/abs/1801.01290}.

\bibitem[Hansen et~al.(1997)Hansen, Ostermeier, and Ostermeier]{Hansen1997}
Nikolaus Hansen, Andreas Ostermeier, and Andreas Ostermeier.
\newblock {Convergence Properties of Evolution Strategies with the Derandomized
  Covariance Matrix Adaptation: CMA-ES}.
\newblock 1997.
\newblock URL
  \url{http://www.cmap.polytechnique.fr/~nikolaus.hansen/CMAES2.pdf}.

\bibitem[Heess et~al.(2015)Heess, Wayne, Silver, Lillicrap, Tassa, and
  Erez]{Heess2015}
Nicolas Heess, Greg Wayne, David Silver, Timothy~P. Lillicrap, Yuval Tassa, and
  Tom Erez.
\newblock Learning continuous control policies by stochastic value gradients.
\newblock \emph{arXiv preprint}, 2015.
\newblock URL \url{http://arxiv.org/abs/1510.09142}.

\bibitem[Hessel et~al.(2018)Hessel, Soyer, Espeholt, Czarnecki, Schmitt, and
  van Hasselt]{Hessel2018}
Matteo Hessel, Hubert Soyer, Lasse Espeholt, Wojciech Czarnecki, Simon Schmitt,
  and Hado van Hasselt.
\newblock {Multi-task Deep Reinforcement Learning with PopArt}.
\newblock \emph{arXiv preprint}, 2018.
\newblock URL \url{https://arxiv.org/pdf/1809.04474.pdf}.

\bibitem[Jaderberg et~al.(2017{\natexlab{a}})Jaderberg, Dalibard, Osindero,
  Czarnecki, Donahue, Razavi, Vinyals, Green, Dunning, Simonyan, Fernando, and
  Kavukcuoglu]{Jaderberg2017}
Max Jaderberg, Valentin Dalibard, Simon Osindero, Wojciech~M. Czarnecki, Jeff
  Donahue, Ali Razavi, Oriol Vinyals, Tim Green, Iain Dunning, Karen Simonyan,
  Chrisantha Fernando, and Koray Kavukcuoglu.
\newblock {Population Based Training of Neural Networks}.
\newblock \emph{arXiv preprint}, 2017{\natexlab{a}}.
\newblock URL \url{http://arxiv.org/abs/1711.09846}.

\bibitem[Jaderberg et~al.(2017{\natexlab{b}})Jaderberg, Mnih, Czarnecki,
  Schaul, Leibo, Silver, and Kavukcuoglu]{Jaderberg2016}
Max Jaderberg, Volodymyr Mnih, Wojciech~Marian Czarnecki, Tom Schaul, Joel~Z
  Leibo, David Silver, and Koray Kavukcuoglu.
\newblock {Reinforcement Learning with Unsupervised Auxiliary Tasks}.
\newblock \emph{Int. Conf. Learn. Represent.}, 2017{\natexlab{b}}.
\newblock URL \url{https://openreview.net/pdf?id=SJ6yPD5xg}.

\bibitem[Jaderberg et~al.(2019)Jaderberg, Czarnecki, Dunning, Marris, Lever,
  Casta{\~n}eda, Beattie, Rabinowitz, Morcos, Ruderman, Sonnerat, Green,
  Deason, Leibo, Silver, Hassabis, Kavukcuoglu, and Graepel]{Jaderberg2019}
Max Jaderberg, Wojciech~M. Czarnecki, Iain Dunning, Luke Marris, Guy Lever,
  Antonio~Garcia Casta{\~n}eda, Charles Beattie, Neil~C. Rabinowitz, Ari~S.
  Morcos, Avraham Ruderman, Nicolas Sonnerat, Tim Green, Louise Deason, Joel~Z.
  Leibo, David Silver, Demis Hassabis, Koray Kavukcuoglu, and Thore Graepel.
\newblock Human-level performance in 3d multiplayer games with population-based
  reinforcement learning.
\newblock \emph{Science}, 364:\penalty0 859--865, 2019.
\newblock URL \url{https://science.sciencemag.org/content/364/6443/859}.

\bibitem[Kapturowski et~al.(2019)Kapturowski, Ostrovski, Quan, Munos, and
  Dabney]{Kapturowski2019}
Steven Kapturowski, Georg Ostrovski, John Quan, R\'{e}mi Munos, and Will
  Dabney.
\newblock {Recurrent Experience Replay in Distributed Reinforcement Learning}.
\newblock \emph{Int. Conf. Learn. Represent.}, 2019.
\newblock URL \url{https://openreview.net/pdf?id=r1lyTjAqYX}.

\bibitem[Kingma \& Ba(2015)Kingma and Ba]{Kingma2015}
Diederik~P. Kingma and Jimmy~Lei Ba.
\newblock {Adam: A method for stochastic optimization}.
\newblock \emph{Int. Conf. Learn. Represent.}, 2015.
\newblock URL \url{https://arxiv.org/abs/1412.6980}.

\bibitem[Levine(2018)]{Levine2018}
Sergey Levine.
\newblock {Reinforcement Learning and Control as Probabilistic Inference:
  Tutorial and Review}.
\newblock \emph{arXiv preprint}, 2018.
\newblock URL \url{http://arxiv.org/abs/1805.00909}.

\bibitem[Lillicrap et~al.(2015)Lillicrap, Hunt, Pritzel, Heess, Erez, Tassa,
  Silver, and Wierstra]{Lillicrap2015}
Timothy~P. Lillicrap, Jonathan~J. Hunt, Alexander Pritzel, Nicolas Heess, Tom
  Erez, Yuval Tassa, David Silver, and Daan Wierstra.
\newblock {Continuous control with deep reinforcement learning}.
\newblock \emph{arXiv preprint}, 2015.
\newblock URL \url{http://arxiv.org/abs/1509.02971}.

\bibitem[Merel et~al.(2018)Merel, Hasenclever, Galashov, Ahuja, Pham, Wayne,
  Teh, and Heess]{Merel2018}
Josh Merel, Leonard Hasenclever, Alexandre Galashov, Arun Ahuja, Vu~Pham, Greg
  Wayne, Yee~Whye Teh, and Nicolas Heess.
\newblock Neural probabilistic motor primitives for humanoid control.
\newblock \emph{arXiv preprint}, 2018.
\newblock URL \url{http://arxiv.org/abs/1811.11711}.

\bibitem[Merel et~al.(2019)Merel, Ahuja, Pham, Tunyasuvunakool, Liu, Tirumala,
  Heess, and Wayne]{Merel2019}
Josh Merel, Arun Ahuja, Vu~Pham, Saran Tunyasuvunakool, Siqi Liu, Dhruva
  Tirumala, Nicolas Heess, and Greg Wayne.
\newblock {Hierarchical Visuomotor Control of Humanoids}.
\newblock \emph{Int. Conf. Learn. Represent.}, 2019.
\newblock URL \url{https://openreview.net/pdf?id=BJfYvo09Y7}.

\bibitem[Mnih et~al.(2015)Mnih, Kavukcuoglu, Silver, Rusu, Veness, Bellemare,
  Graves, Riedmiller, Fidjeland, Ostrovski, Petersen, Beattie, Sadik,
  Antonoglou, King, Kumaran, Wierstra, Legg, and Hassabis]{Mnih2015}
Volodymyr Mnih, Koray Kavukcuoglu, David Silver, Andrei~A Rusu, Joel Veness,
  Marc~G Bellemare, Alex Graves, Martin Riedmiller, Andreas~K Fidjeland, Georg
  Ostrovski, Stig Petersen, Charles Beattie, Amir Sadik, Ioannis Antonoglou,
  Helen King, Dharshan Kumaran, Daan Wierstra, Shane Legg, and Demis Hassabis.
\newblock {Human-Level Control through Deep Reinforcement Learning}.
\newblock \emph{Nature}, 518:\penalty0 529--533, 2015.
\newblock URL \url{http://dx.doi.org/10.1038/nature14236}.

\bibitem[Mnih et~al.(2016)Mnih, Badia, Mirza, Graves, Harley, Lillicrap,
  Silver, and Kavukcuoglu]{Mnih2016}
Volodymyr Mnih, Adri\`{a}~Puigdom\`{e}nech Badia, Mehdi Mirza, Alex Graves, Tim
  Harley, Timothy~P Lillicrap, David Silver, and Koray Kavukcuoglu.
\newblock {Asynchronous Methods for Deep Reinforcement Learning}.
\newblock \emph{arXiv:1602.01783}, 2016.
\newblock URL \url{http://arxiv.org/abs/1602.01783}.

\bibitem[Neal \& Hinton(1998)Neal and Hinton]{Neal1998}
Radford~M. Neal and Geoffrey~E. Hinton.
\newblock {A View of the {EM} Algorithm that Justifies Incremental, Sparse, and
  Other Variants}.
\newblock In M.I. Jordan (ed.), \emph{Learn. Graph. Model. NATO ASI Ser. vol.
  89}. Springer, Dordrecht, 1998.

\bibitem[{OpenAI}(2018{\natexlab{a}})]{OpenAI2018}
{OpenAI}.
\newblock {OpenAI Five}, 2018{\natexlab{a}}.
\newblock URL \url{https://openai.com/blog/openai-five/}.

\bibitem[{OpenAI}(2018{\natexlab{b}})]{OpenAI2018a}
{OpenAI}.
\newblock {Learning Dexterity}, 2018{\natexlab{b}}.
\newblock URL \url{https://openai.com/blog/learning-dexterity/}.

\bibitem[Peters et~al.(2008)Peters, Katharina, and Alt\"{u}n]{Peters2008a}
Jan Peters, M~Katharina, and Yasemin Alt\"{u}n.
\newblock {Relative Entropy Policy Search}.
\newblock \emph{Proceedings of the Twenty-Fourth AAAI Conference on Artificial
  Intelligence}, pp.\  1607--1612, 2008.

\bibitem[Radford et~al.(2019)Radford, Wu, Child, Luan, Amodei, and
  Sutskever]{Radford2019}
Alec Radford, Jeff Wu, Rewon Child, David Luan, Dario Amodei, and Ilya
  Sutskever.
\newblock {Language Models are Unsupervised Multitask Learners}.
\newblock 2019.
\newblock URL
  \url{https://d4mucfpksywv.cloudfront.net/better-language-models/language_models_are_unsupervised_multitask_learners.pdf}.

\bibitem[Schmitt et~al.(2018)Schmitt, Hudson, Z{\'{\i}}dek, Osindero, Doersch,
  Czarnecki, Leibo, K{\"{u}}ttler, Zisserman, Simonyan, and
  Eslami]{Schmitt2018}
Simon Schmitt, Jonathan~J. Hudson, Augustin Z{\'{\i}}dek, Simon Osindero, Carl
  Doersch, Wojciech~M. Czarnecki, Joel~Z. Leibo, Heinrich K{\"{u}}ttler, Andrew
  Zisserman, Karen Simonyan, and S.~M.~Ali Eslami.
\newblock {Kickstarting Deep Reinforcement Learning}.
\newblock \emph{arXiv preprint}, 2018.
\newblock URL \url{http://arxiv.org/abs/1803.03835}.

\bibitem[Schulman et~al.(2015)Schulman, Levine, Moritz, Jordan, and
  Abbeel]{Schulman2015a}
John Schulman, Sergey Levine, Philipp Moritz, Michael~I. Jordan, and Pieter
  Abbeel.
\newblock {Trust Region Policy Optimization}.
\newblock \emph{arXiv preprint}, 2015.
\newblock URL \url{http://arxiv.org/abs/1502.05477}.

\bibitem[Schulman et~al.(2017)Schulman, Wolski, Dhariwal, Radford, and
  Klimov]{Schulman2017}
John Schulman, Filip Wolski, Prafulla Dhariwal, Alec Radford, and Oleg Klimov.
\newblock Proximal policy optimization algorithms.
\newblock \emph{arXiv preprint}, 2017.
\newblock URL \url{http://arxiv.org/abs/1707.06347}.

\bibitem[Silver et~al.(2016)Silver, Huang, Maddison, Guez, Sifre, van~den
  Driessche, Schrittwieser, Antonoglou, Panneershelvam, Lanctot, Dieleman,
  Grewe, Nham, Kalchbrenner, Sutskever, Lillicrap, Leach, Kavukcuoglu, Graepel,
  and Hassabis]{Silver2016}
David Silver, Aja Huang, Chris~J. Maddison, Arthur Guez, Laurent Sifre, George
  van~den Driessche, Julian Schrittwieser, Ioannis Antonoglou, Veda
  Panneershelvam, Marc Lanctot, Sander Dieleman, Dominik Grewe, John Nham, Nal
  Kalchbrenner, Ilya Sutskever, Timothy Lillicrap, Madeleine Leach, Koray
  Kavukcuoglu, Thore Graepel, and Demis Hassabis.
\newblock {Mastering the game of Go with deep neural networks and tree search}.
\newblock \emph{Nature}, 529:\penalty0 484--489, 2016.
\newblock URL \url{http://www.nature.com/doifinder/10.1038/nature16961}.

\bibitem[Silver et~al.(2018)Silver, Hubert, Schrittwieser, Antonoglou, Lai,
  Guez, Lanctot, Sifre, Kumaran, Graepel, Lillicrap, Simonyan, and
  Hassabis]{Silver2018}
David Silver, Thomas Hubert, Julian Schrittwieser, Ioannis Antonoglou, Matthew
  Lai, Arthur Guez, Marc Lanctot, Laurent Sifre, Dharshan Kumaran, Thore
  Graepel, Timothy Lillicrap, Karen Simonyan, and Demis Hassabis.
\newblock A general reinforcement learning algorithm that masters chess, shogi,
  and go through self-play.
\newblock \emph{Science}, 362:\penalty0 1140--1144, 2018.
\newblock URL \url{https://science.sciencemag.org/content/362/6419/1140}.

\bibitem[Sutton \& Barto(1998)Sutton and Barto]{Sutton1998}
Richard~S. Sutton and Andrew~G. Barto.
\newblock \emph{{Reinforcement Learning: An Introduction}}.
\newblock MIT Press, Cambridge, MA, 1998.

\bibitem[Sutton et~al.(2000)Sutton, McAllester, Singh, and Mansour]{Sutton2000}
Richard~S Sutton, David~A. McAllester, Satinder~P. Singh, and Yishay Mansour.
\newblock Policy gradient methods for reinforcement learning with function
  approximation.
\newblock In S.~A. Solla, T.~K. Leen, and K.~M\"{u}ller (eds.), \emph{Advances
  in Neural Information Processing Systems 12}, pp.\  1057--1063. MIT Press,
  2000.
\newblock URL
  \url{http://papers.nips.cc/paper/1713-policy-gradient-methods-for-reinforcement-learning-with-function-approximation.pdf}.

\bibitem[Tassa et~al.(2018)Tassa, Doron, Muldal, Erez, Li, de~Las~Casas,
  Budden, Abdolmaleki, Merel, Lefrancq, Lillicrap, and Riedmiller]{Tassa2018}
Yuval Tassa, Yotam Doron, Alistair Muldal, Tom Erez, Yazhe Li, Diego
  de~Las~Casas, David Budden, Abbas Abdolmaleki, Josh Merel, Andrew Lefrancq,
  Timothy~P. Lillicrap, and Martin~A. Riedmiller.
\newblock {DeepMind Control Suite}.
\newblock \emph{arXiv preprint}, 2018.
\newblock URL \url{http://arxiv.org/abs/1801.00690}.

\bibitem[van Hasselt et~al.(2016)van Hasselt, Guez, Hessel, and
  Silver]{Hasselt2016}
Hado van Hasselt, Arthur Guez, Matteo Hessel, and David Silver.
\newblock Learning functions across many orders of magnitudes.
\newblock \emph{arXiv preprint}, 2016.
\newblock URL \url{http://arxiv.org/abs/1602.07714}.

\bibitem[Vuong et~al.(2019)Vuong, Ross, and Zhang]{Vuong2018}
Quan Vuong, Keith Ross, and Yiming Zhang.
\newblock {Supervised Policy Update for Deep Reinforcement Learning}.
\newblock \emph{arXiv preprint}, 2019.
\newblock URL \url{http://arxiv.org/abs/1805.11706}.

\bibitem[Williams(1992)]{Williams1992}
Ronald~J. Williams.
\newblock {Simple statistical gradient-following methods for connectionist
  reinforcement learning}.
\newblock \emph{Mach. Learn.}, 8:\penalty0 229--256, 1992.
\newblock URL \url{http://dx.doi.org/10.1007/BF00992696}.

\bibitem[Wu et~al.(2018)Wu, Rajeswaran, Duan, Kumar, Bayen, Kakade, Mordatch,
  and Abbeel]{Wu2018}
Cathy Wu, Aravind Rajeswaran, Yan Duan, Vikash Kumar, Alexandre~M. Bayen, Sham
  Kakade, Igor Mordatch, and Pieter Abbeel.
\newblock Variance reduction for policy gradient with action-dependent
  factorized baselines.
\newblock \emph{arXiv preprint}, 2018.
\newblock URL \url{http://arxiv.org/abs/1803.07246}.

\end{thebibliography}
\bibliographystyle{iclr2020_conference}


\appendix


\section{Derivation of the V-MPO temperature loss}
\label{appendix:temperature_loss}

In this section we derive the E-step temperature loss in Eq.~\ref{eq:full_temperature_loss}. To this end, we explicitly commit to the more specific improvement criterion in Eq.~\ref{eq:improvement_criterion} by plugging into the original objective in Eq.~\ref{eq:elbo}. We seek $\psi(s,a)$ that minimizes
\begin{align}
    \mathcal{J}(\psi(s,a))
    &= D_\text{KL}\big( \psi(s,a) \| p_{\theta_\text{old}}(s,a|\mathcal{I}=1) \big) \\
    &\propto -\sum_{s,a} \psi(s,a) A^{\pi_{\theta_\text{old}}}(s,a) 
    + \eta \sum_{s,a} \psi(s,a) \log \frac{\psi(s,a)}{p_{\theta_\text{old}}(s,a)}
    + \lambda \sum_{s,a} \psi(s,a) 
    \label{eq:e_step_kl}
\end{align} where $\lambda=\eta \log p_{\theta_\text{old}}(\mathcal{I}=1)$ after multiplying through by $\eta$, which up to this point in the derivation is given. We wish to automatically tune $\eta$ so as to enforce a bound $\epsilon_\eta$ on the KL term $D_\text{KL}\big( \psi(s,a) \| p_{\theta_\text{old}}(s,a) \big)$ multiplying it in Eq.~\ref{eq:e_step_kl}, in which case the temperature optimization can also be viewed as a nonparametric trust region for the variational distribution with respect to the old distribution. We therefore consider the constrained optimization problem
\begin{align}
    &\quad\psi(s,a) = \argmax{\psi(s,a)} \sum_{s,a} \psi(s,a) A^{\pi_{\theta_\text{old}}}(s,a) \\
    \text{s.t. } & \sum_{s,a} \psi(s,a) \log \frac{\psi(s,a)}{p_{\theta_\text{old}}(s,a)} < \epsilon_\eta \text{ and } \sum_{s,a} \psi(s,a) = 1. \label{eq:e_step_constraint}
\end{align} We can now use Lagrangian relaxation to transform the constrained optimization problem into one that maximizes the unconstrained objective
\begin{equation}
    \mathcal{J}(\psi(s,a), \eta,\lambda) = 
    \sum_{s,a} \psi(s,a) A^{\pi_{\theta_\text{old}}}(s,a) + \eta\Bigg( \epsilon_\eta - \sum_{s,a} \psi(s,a) \log \frac{\psi(s,a)}{p_{\theta_\text{old}}(s,a)} \Bigg) + \lambda\Bigg(1 - \sum_{s,a} \psi(s,a) \Bigg) \label{eq:unconstrained_objective}
\end{equation} with $\eta \geq0$. (Note we are re-using the variables $\eta$ and $\lambda$ for the new optimization problem.) Differentiating $\mathcal{J}$ with respect to $\psi(s,a)$ and setting equal to zero, we obtain
\begin{equation}
    \psi(s,a) = p_{\theta_\text{old}}(s,a) \exp \bigg( \frac{A^{\pi_{\theta_\text{old}}}(s,a)}{\eta} \bigg) \exp \bigg( -1 - \frac{\lambda}{\eta} \bigg). \label{eq:mu_plus_from_derivative}
\end{equation} Normalizing over $s,a$ (using the freedom given by $\lambda$) then gives
\begin{equation}
    \psi(s,a) = \frac{p_{\theta_\text{old}}(s,a) \exp \big( \frac{A^{\pi_{\theta_\text{old}}}(s,a)}{\eta} \big)}{\sum_{s,a} p_{\theta_\text{old}}(s,a) \exp \big( \frac{A^{\pi_{\theta_\text{old}}}(s,a)}{\eta} \big)}, \label{eq:mu_plus_normalized}
\end{equation} which reproduces the general solution Eq.~\ref{eq:psi_general} for our specific choice of policy improvement in Eq.~\ref{eq:improvement_criterion}. However, the value of $\eta$ can now be found by optimizing the corresponding dual function. Plugging Eq.~\ref{eq:mu_plus_normalized} into the unconstrained objective in Eq.~\ref{eq:unconstrained_objective} gives rise to the $\eta$-dependent term
\begin{equation}
    \mathcal{L}_\eta(\eta) = \eta \epsilon_\eta + \eta \log \Bigg[ \sum_{s,a} p_{\theta_\text{old}}(s,a) \exp \bigg( \frac{A^{\pi_{\theta_\text{old}}}(s,a)}{\eta} \bigg) \Bigg]. \label{eq:full_temperature_loss}
\end{equation} Replacing the expectation with samples from $p_{\theta_\text{old}}(s,a)$ in the batch of trajectories $\mathcal{D}$ leads to the loss in Eq.~\ref{eq:L_eta}.


\section{M-step KL constraint}
\label{appendix:kl-constraint}

Here we give a somewhat more formal motivation for the prior $\log p(\theta)$. Consider a normal prior $\mathcal{N}(\theta;\mu,\Sigma)$ with mean $\mu$ and covariance $\Sigma$. We choose $\Sigma^{-1}=\alpha F(\theta_\text{old})$ where $\alpha$ is a scaling parameter and $F(\theta_\text{old})$ is the Fisher information for $\pi_{\theta'}(a|s)$ evaluated at $\theta'=\theta_\text{old}$. Then $\log p(\theta) \approx -\alpha \times \frac{1}{2}(\theta - \theta_\text{old})^T F(\theta_\text{old}) (\theta - \theta_\text{old}) + \{ \text{term independent of $\theta$} \}$, where the first term is precisely the second-order approximation to the KL divergence $\dkl(\theta_\text{old} \| \vphantom{\theta_\text{old}}\theta )$. We now follow TRPO~\cite{Schulman2015a} in heuristically approximating this as the state-averaged expression, $\mathbb{E}_{s\sim p(s)} \big[ \dkl\big( \pi_{\theta_\text{old}}(a|s) \| \pi_{\vphantom{\theta_\text{old}}\theta}(a|s) \big) \big]$. We note that the KL divergence in either direction has the same second-order expansion, so our choice of KL is an empirical one~\cite{Abdolmaleki2018a}.


\section{Decoupled KL constraints for continuous control}
\label{appendix:continuous_control}

As in \citeauthor{Abdolmaleki2018}~(\citeyear{Abdolmaleki2018}), for continuous action spaces parametrized by Gaussian distributions we use decoupled KL constraints for the M-step. This uses the fact that the KL divergence between two $d$-dimensional multivariate normal distributions with means $\mu_1,\mu_2$ and covariances $\Sigma_1,\Sigma_2$ can be written as
\begin{equation}
    D_\text{KL}\big( \mathcal{N}(\mu_1,\Sigma_1) \| \mathcal{N}(\mu_2,\Sigma_2) \big)  = \frac{1}{2}\bigg[
    (\mu_2 - \mu_1)^T\Sigma_1^{-1}(\mu_2 - \mu_1) + \operatorname{Tr}(\Sigma_2^{-1}\Sigma_1) - d + \log \frac{|\Sigma_2|}{|\Sigma_1|}
    \bigg],
\end{equation} where $|\cdot|$ is the matrix determinant. Since the first distribution and hence $\Sigma_1$ in the KL divergence of Eq.~\ref{eq:m_step_objective} depends on the old target network parameters, we see that we can separate the overall KL divergence into a mean component and a covariance component:
\begin{align}
    D^\mu_\text{KL}\big( \pi_{\theta_\text{old}} \| \pi_{\vphantom{\theta_\text{old}}\theta} \big) &= \frac{1}{2}(\mu_\theta - \mu_{\theta_\text{old}})^T\Sigma_{\theta_\text{old}}^{-1}(\mu_\theta - \mu_{\theta_\text{old}}),\\
    D^\Sigma_\text{KL}\big( \pi_{\theta_\text{old}} \| \pi_{\vphantom{\theta_\text{old}}\theta} \big) &= \frac{1}{2}\bigg[ \operatorname{Tr}(\Sigma_{\vphantom{\theta_\text{old}}\theta}^{-1}\Sigma^{\vphantom{-1}}_{\theta_\text{old}}) - d + \log \frac{| \Sigma_\theta|}{|\Sigma_{\theta_\text{old}}|} \bigg].
\end{align} With the replacement $D_\text{KL}\big( \pi_{\theta_\text{old}} \| \pi_{\vphantom{\theta_\text{old}}\theta} \big)\rightarrow D^C_\text{KL}\big( \pi_{\theta_\text{old}} \| \pi_{\vphantom{\theta_\text{old}}\theta} \big)$ for $C=\mu,\Sigma$ and corresponding $\alpha\rightarrow \alpha_\mu,\ \alpha_\Sigma$ in Eq.~\ref{eq:m_step_kl_constraint}, we obtain the total loss
\begin{equation}
    \mathcal{L}_\text{V-MPO}(\theta, \eta, \alpha_\mu, \alpha_\Sigma) = \mathcal{L}_\pi(\theta) + \mathcal{L}_\eta(\eta) + \mathcal{L}_{\alpha_\mu}(\theta,\alpha_\mu) + \mathcal{L}_{\alpha_\Sigma}(\theta,\alpha_\Sigma),
\end{equation} where $\mathcal{L}_\pi(\theta)$ and $\mathcal{L}_\eta(\eta)$ are the same as before. Note, however, that unlike in \citeauthor{Abdolmaleki2018a}~(\citeyear{Abdolmaleki2018a}) we do not decouple the policy loss.

We generally set $\epsilon_\Sigma$ to be much smaller than $\epsilon_\mu$ (see Table~\ref{table:continuous_control_settings}). Intuitively, this allows the policy to learn quickly in action space while preventing premature collapse of the policy, and, conversely, increasing ``exploration'' without moving in action space.


\section{Relation to Supervised Policy Update}
\label{appendix:spu}

Like V-MPO, Supervised Policy Update (SPU)~\cite{Vuong2018} adopts the strategy of first solving a nonparametric constrained optimization problem exactly, then fitting a neural network to the resulting solution via a supervised loss function. There is, however, an important difference from V-MPO, which we describe here.

In SPU, the KL loss, which is the \emph{sole} loss in SPU, leads to a parametric optimization problem that is equivalent to the nonparametric optimization problem posed initially. To see this, we observe that the SPU loss seeks parameters (note the direction of the KL divergence)
\begin{align}
    \theta^* &= \argmin{\theta} \sum_s d^{\pi_{\theta_k}}(s) D_\text{KL}\big( \pi_\theta(a|s) \| \pi^\lambda(a|s) \big) \\
    &= \argmin{\theta} \sum_s d^{\pi_{\theta_k}}(s) \sum_a \pi_\theta(a|s) \log \bigg[ \frac{\pi_\theta(a|s)}{\pi_{\theta_k}(a|s)\exp\big( A^{\pi_{\theta_k}}(s,a)/\lambda\big) / Z_\lambda(s)} \bigg] \\
    &= \argmin{\theta} \sum_s d^{\pi_{\theta_k}}(s) \sum_a \bigg[ \pi_\theta(a|s) \log \frac{\pi_\theta(a|s)}{\pi_{\theta_k}(a|s)} - \frac{1}{\lambda}\pi_\theta(a|s) A^{\pi_{\theta_k}}(s,a) \bigg] + \{\text{ constant terms } \}.
\end{align} Multiplying by $\lambda$ since it can be treated as a constant up to this point, we then see that this corresponds exactly to the (Lagrangian form) of the problem
\begin{align}
    &\theta^* = \argmax{\theta} \sum_s d^{\pi_{\theta_k}}(s) \sum_a \pi_\theta(a|s) A^{\pi_{\theta_k}}(s,a) \\
    &\qquad \text{s.t.} \sum_s d^{\pi_{\theta_k}}(s) D_\text{KL}\big( \pi_{\vphantom{\theta_k}\theta}(a|s) \| \pi_{\theta_k}(a|s) \big) < \epsilon,
\end{align} which is the original nonparametric problem posed in \citeauthor{Vuong2018}~(\citeyear{Vuong2018}).


\section{Importance-weighting for off-policy corrections}
\label{appendix:importance_weighting}

The network that generates the data may lag behind the target network in common distributed, asynchronous implementations~\cite{Espeholt2018}. We can compensate for this by multiplying the exponentiated advantages by importance weights $\rho(s,a)$:
\begin{align}
    \psi(s,a) &= \frac{\rho(s,a)p_{\theta_\mathcal{D}}(s,a) \exp \big( \frac{A^{\pi_{\theta_\mathcal{D}}}(s,a)}{\eta} \big)}{\sum_{s,a} \rho(s,a) p_{\theta_\mathcal{D}}(s,a) \exp \big( \frac{A^{\pi_{\theta_\mathcal{D}}}(s,a)}{\eta} \big)}, \\
    \mathcal{L}_\eta(\eta) &= \eta \epsilon_\eta + \eta \log \Bigg[ \sum_{s,a} \rho(s,a)p_{\theta_\mathcal{D}}(s,a) \exp \bigg( \frac{A^{\pi_{\theta_\mathcal{D}}}(s,a)}{\eta} \bigg) \Bigg],
\end{align} where $\theta_\mathcal{D}$ are the parameters of the behavior policy that generated $\mathcal{D}$ and which may be different from $\theta_\text{target}$. The clipped importance weights $\rho(s,a)$ are given by
\begin{equation}
    \rho(s,a) = \operatorname{min}\bigg( 1, \frac{\pi_{\theta_\text{old}}(a|s)}{\pi_{\theta_\mathcal{D}}(a|s)} \bigg).
\end{equation}

As was the case with V-trace for the value function, we did not find it necessary to use importance weighting and all experiments presented in this work did not use them for the sake of simplicity.


\section{Network architecture and hyperparameters}
\label{appendix:network_architecture}

\begin{figure}[t!]
\begin{center}
\includegraphics[width=0.75\linewidth]{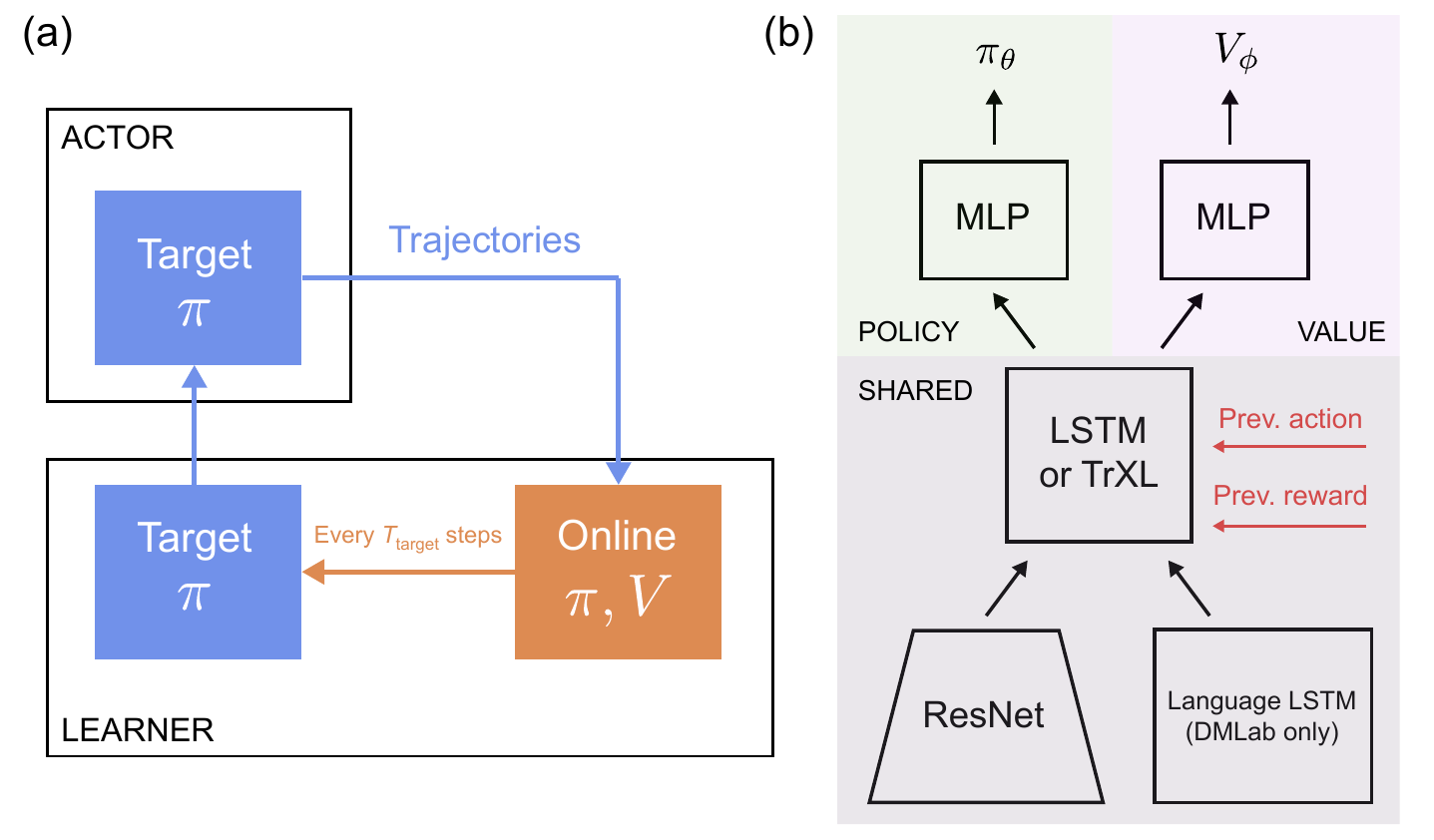}
\caption{(a) Actor-learner architecture with a target network, which is used to generate agent experience in the environment and is updated every $T_\text{target}$ learning steps from the online network. (b) Schematic of the agents, with the policy ($\theta$) and value ($\phi$) networks sharing most of their parameters through a shared input encoder and LSTM [or Transformer-XL (TrXL) for single Atari levels]. The agent also receives the action and reward from the previous step as an input to the LSTM. For DMLab an additional LSTM is used to process simple language instructions.}
\label{fig:schematic}
\end{center}
\end{figure}

For DMLab the visual observations were 72$\times$96 RGB images, while for Atari the observations were 4 stacked frames of 84$\times$84 grayscale images. The ResNet used to process visual observations is similar to the 3-section ResNet used in \citeauthor{Hessel2018}~(\citeyear{Hessel2018}), except the number of channels was multiplied by 4 in each section, so that the number of channels were (64, 128, 128)~\cite{Schmitt2019}. For individual DMLab levels we used the same number of channels as \citeauthor{Hessel2018}~(\citeyear{Hessel2018}), i.e., (16, 32, 32). Each section consisted of a convolution and $3\times3$ max-pooling operation (stride 2), followed by residual blocks of size 2, i.e., a convolution followed by a ReLU nonlinearity, repeated twice, and a skip connection from the input residual block input to the output. The entire stack was passed through one more ReLU nonlinearity. All convolutions had a kernel size of 3 and a stride of 1. For the humanoid control tasks from vision, the number of channels in each section were (16, 32, 32). 

Since some of the levels in DMLab require simple language processing, for DMLab the agents contained an additional 256-unit LSTM receiving an embedding of hashed words as input. The output of the language LSTM was then concatenated with the output of the visual processing pathway as well as the previous reward and action, then fed to the main LSTM.

For multi-task DMLab we used a 3-layer LSTM, each with 256 units, and an unroll length of 95 with batch size 128. For the single-task setting we used a 2-layer LSTM. For multi-task Atari and the 56-dimensional humanoid-gaps control task a single 256-unit LSTM was used, while for the 22-dimensional humanoid-run task the core consisted only of a 2-layer MLP with 512 and 256 units (no LSTM). For single-task Atari a Transformer-XL was used in place of the LSTM. Note that we followed \citeauthor{Radford2019}~(\citeyear{Radford2019}) in placing the layer normalization on only the inputs to each sub-block. For Atari the unroll length was 63 with a batch size of 128. For both humanoid control tasks the batch size was 64, but the unroll length was 40 for the 22-dimensional humanoid and 63 for the 56-dimensional humanoid.

In all cases the policy logits (for discrete actions) and Gaussian distribution parameters (for continuous actions) consisted of a 256-unit MLP followed by a linear readout, and similarly for the value function.

The initial values for the Lagrange multipliers in the V-MPO loss are given in Table~\ref{table:initial_lagrange_multipliers}

\textit{Implementation note.} We implemented V-MPO in an actor-learner framework~\cite{Espeholt2018} that utilizes TF-Replicator~\cite{Buchlovsky2019} for distributed training on TPU 8-core and 16-core configurations~\cite{Google2018}. One practical consequence of this is that a full batch of data $\mathcal{D}$ was in fact split into 8 or 16 minibatches, one per core/replica, and the overall result obtained by averaging the computations performed for each minibatch. More specifically, the determination of the highest advantages and the normalization of the nonparametric distribution, Eq.~\ref{eq:policy_loss}, is performed within minibatches. While it is possible to perform the full-batch computation by utilizing cross-replica communication, we found this to be unnecessary.


\begin{table}[h!]
\begin{center}
\begin{small}
\begin{tabular}{lccc}
\toprule
{\sc Hyperparameter} &  \multicolumn{3}{c}{{\sc Value}} \\[0.5em]
& DMLab & Atari & Continuous control \\
\midrule

Initial $\eta$ & 1.0 & 1.0 & 1.0 \\
Initial $\alpha$ & 5.0 & 5.0 & - \\
Initial $\alpha_\mu$ & - & - & 1.0 \\
Initial $\alpha_\Sigma$ & - & - & 1.0 \\

\bottomrule
\end{tabular}
\end{small}

\caption{Values for common V-MPO parameters.}
\label{table:initial_lagrange_multipliers}

\end{center}
\end{table}

\emph{DMLab action set.} Ignoring the ``jump'' and ``crouch'' actions which we do not use, an action in the native DMLab action space consists of 5 integers whose meaning and allowed values are given in Table~\ref{table:dmlab_native_action_set}. Following previous work on DMLab~\cite{Hessel2018}, we used the reduced action set given in Table~\ref{table:dmlab_action_set} with an action repeat of 4.


\begin{table}[h!]
\begin{center}
\begin{small}
\begin{tabular}{lc}
\toprule
{\sc Action name} & {\sc Range} \\
\midrule

LOOK\_LEFT\_RIGHT\_PIXELS\_PER\_FRAME & [-512, 512] \\
LOOK\_DOWN\_UP\_PIXELS\_PER\_FRAME & [-512, 512] \\
STRAFE\_LEFT\_RIGHT & [-1, 1] \\
MOVE\_BACK\_FORWARD & [-1, 1] \\
FIRE & [0, 1] \\

\bottomrule
\end{tabular}
\end{small}

\caption{Native action space for DMLab. See \url{https://github.com/deepmind/lab/blob/master/docs/users/actions.md} for more details.}
\label{table:dmlab_native_action_set}

\end{center}
\end{table}


\begin{table}[h!]
\begin{center}
\begin{small}
\begin{tabular}{lc}
\toprule
{\sc Action} & {\sc Native DMLab action} \\
\midrule

Forward (FW) & \texttt{[~~0,~~~0,~~0,~~1,~0]}\\
Backward (BW) & \texttt{[~~0,~~~0,~~0,~-1,~0]} \\[0.5em]

Strafe left & \texttt{[~~0,~~~0,~-1,~~0,~0]} \\
Strafe right & \texttt{[~~0,~~~0,~~1,~~0,~0]} \\[0.5em]

Small look left (LL) & \texttt{[-10,~~~0,~~0,~~0,~0]} \\
Small look right (LR) & \texttt{[~10,~~~0,~~0,~~0,~0]} \\
Large look left (LL ) & \texttt{[-60,~~~0,~~0,~~0,~0]} \\
Large look right (LR) & \texttt{[ 60,~~~0,~~0,~~0,~0]} \\[0.5em]

Look down & \texttt{[~~0,~~10,~~0,~~0,~0]} \\
Look up & \texttt{[~~0,~-10,~~0,~~0,~0]} \\[0.5em]

FW + small LL & \texttt{[-10,~~~0,~~0,~~1,~0]} \\
FW + small LR & \texttt{[ 10,~~~0,~~0,~~1,~0]} \\
FW + large LL & \texttt{[-60,~~~0,~~0,~~1,~0]} \\
FW + large LR & \texttt{[ 60,~~~0,~~0,~~1,~0]} \\[0.5em]

Fire & \texttt{[~~0,~~~0,~~0,~~0,~1]} \\

\bottomrule
\end{tabular}
\end{small}

\caption{Reduced action set for DMLab from \citeauthor{Hessel2018}~(\citeyear{Hessel2018}).}
\label{table:dmlab_action_set}

\end{center}
\end{table}


\begin{table}[t!]
\begin{center}
\begin{footnotesize}
\begin{tabular}{l|cc|cc}
\toprule
{\sc Level name} & \multicolumn{2}{c}{{\sc Episode reward}} & \multicolumn{2}{c}{{\sc Human-normalized}} \\
& IMPALA & V-MPO & IMPALA & V-MPO \\
\midrule

alien               &    1163.00 $\pm$     148.43 &    2332.00 $\pm$     290.16&      13.55 $\pm$       2.15 &      30.50 $\pm$       4.21 \\
amidar              &     192.50 $\pm$       9.16 &     423.60 $\pm$      20.53&      10.89 $\pm$       0.53 &      24.38 $\pm$       1.20 \\
assault             &    4215.30 $\pm$     294.51 &    1225.90 $\pm$      60.64&     768.46 $\pm$      56.68 &     193.13 $\pm$      11.67 \\
asterix             &    4180.00 $\pm$     303.91 &    9955.00 $\pm$    2043.48&      47.87 $\pm$       3.66 &     117.50 $\pm$      24.64 \\
asteroids           &    3473.00 $\pm$     381.30 &    2982.00 $\pm$     164.35&       5.90 $\pm$       0.82 &       4.85 $\pm$       0.35 \\
atlantis            &  997530.00 $\pm$    3552.89 &  940310.00 $\pm$    6085.96&    6086.50 $\pm$      21.96 &    5732.81 $\pm$      37.62 \\
bank\_heist         &    1329.00 $\pm$       2.21 &    1563.00 $\pm$      15.81&     177.94 $\pm$       0.30 &     209.61 $\pm$       2.14 \\
battle\_zone        &   43900.00 $\pm$    4738.04 &   61400.00 $\pm$    5958.52&     119.27 $\pm$      13.60 &     169.52 $\pm$      17.11 \\
beam\_rider         &    4598.00 $\pm$     618.09 &    3868.20 $\pm$     666.55&      25.56 $\pm$       3.73 &      21.16 $\pm$       4.02 \\
berzerk             &    1018.00 $\pm$      72.63 &    1424.00 $\pm$     150.93&      35.68 $\pm$       2.90 &      51.87 $\pm$       6.02 \\
bowling             &      63.60 $\pm$       0.84 &      27.60 $\pm$       0.62&      29.43 $\pm$       0.61 &       3.27 $\pm$       0.45 \\
boxing              &      93.10 $\pm$       0.94 &     100.00 $\pm$       0.00&     775.00 $\pm$       7.86 &     832.50 $\pm$       0.00 \\
breakout            &     484.30 $\pm$      57.24 &     400.70 $\pm$      18.82&    1675.69 $\pm$     198.77 &    1385.42 $\pm$      65.36 \\
centipede           &    6037.90 $\pm$     994.99 &    3015.00 $\pm$     404.97&      39.76 $\pm$      10.02 &       9.31 $\pm$       4.08 \\
chopper\_command    &    4250.00 $\pm$     417.91 &    4340.00 $\pm$     714.45&      52.29 $\pm$       6.35 &      53.66 $\pm$      10.86 \\
crazy\_climber      &  100440.00 $\pm$    9421.56 &  116760.00 $\pm$    5312.12&     357.94 $\pm$      37.61 &     423.09 $\pm$      21.21 \\
defender            &   41585.00 $\pm$    4194.42 &   98395.00 $\pm$   17552.17&     244.78 $\pm$      26.52 &     604.01 $\pm$     110.99 \\
demon\_attack       &   77880.00 $\pm$    8798.44 &   20243.00 $\pm$    5434.41&    4273.35 $\pm$     483.72 &    1104.56 $\pm$     298.77 \\
double\_dunk        &      -0.80 $\pm$       0.31 &      12.60 $\pm$       1.94&     809.09 $\pm$      14.08 &    1418.18 $\pm$      88.19 \\
enduro              &    1187.90 $\pm$      76.10 &    1453.80 $\pm$     104.37&     138.05 $\pm$       8.84 &     168.95 $\pm$      12.13 \\
fishing\_derby      &      21.60 $\pm$       3.46 &      33.80 $\pm$       2.10&     213.77 $\pm$       6.54 &     236.79 $\pm$       3.96 \\
freeway             &      32.10 $\pm$       0.17 &      33.20 $\pm$       0.28&     108.45 $\pm$       0.58 &     112.16 $\pm$       0.93 \\
frostbite           &     250.00 $\pm$       0.00 &     260.00 $\pm$       0.00&       4.33 $\pm$       0.00 &       4.56 $\pm$       0.00 \\
gopher              &   11720.00 $\pm$    1687.71 &    7576.00 $\pm$     973.13&     531.92 $\pm$      78.32 &     339.62 $\pm$      45.16 \\
gravitar            &    1095.00 $\pm$     232.75 &    3125.00 $\pm$     191.87&      29.01 $\pm$       7.32 &      92.88 $\pm$       6.04 \\
hero                &   13159.50 $\pm$      68.90 &   29196.50 $\pm$     752.06&      40.71 $\pm$       0.23 &      94.53 $\pm$       2.52 \\
ice\_hockey         &       4.80 $\pm$       1.31 &      10.60 $\pm$       2.00&     132.23 $\pm$      10.83 &     180.17 $\pm$      16.50 \\
jamesbond           &    1015.00 $\pm$      91.39 &    3805.00 $\pm$     595.92&     360.12 $\pm$      33.38 &    1379.11 $\pm$     217.65 \\
kangaroo            &    1780.00 $\pm$      18.97 &   12790.00 $\pm$     629.52&      57.93 $\pm$       0.64 &     427.02 $\pm$      21.10 \\
krull               &    9738.00 $\pm$     360.95 &    7359.00 $\pm$    1064.84&     762.53 $\pm$      33.81 &     539.67 $\pm$      99.75 \\
kung\_fu\_master    &   44340.00 $\pm$    2898.70 &   38620.00 $\pm$    2346.48&     196.11 $\pm$      12.90 &     170.66 $\pm$      10.44 \\
montezuma\_revenge  &       0.00 $\pm$       0.00 &       0.00 $\pm$       0.00&       0.00 $\pm$       0.00 &       0.00 $\pm$       0.00 \\
ms\_pacman          &    1953.00 $\pm$     227.12 &    2856.00 $\pm$     324.54&      24.77 $\pm$       3.42 &      38.36 $\pm$       4.88 \\
name\_this\_game    &    5708.00 $\pm$     354.92 &    9295.00 $\pm$     679.83&      59.33 $\pm$       6.17 &     121.64 $\pm$      11.81 \\
phoenix             &   37030.00 $\pm$    6415.95 &   19560.00 $\pm$    1843.44&     559.60 $\pm$      98.99 &     290.05 $\pm$      28.44 \\
pitfall             &      -4.90 $\pm$       2.34 &      -2.80 $\pm$       1.40&       3.35 $\pm$       0.04 &       3.39 $\pm$       0.02 \\
pong                &      20.80 $\pm$       0.19 &      21.00 $\pm$       0.00&     117.56 $\pm$       0.54 &     118.13 $\pm$       0.00 \\
private\_eye        &     100.00 $\pm$       0.00 &     100.00 $\pm$       0.00&       0.11 $\pm$       0.00 &       0.11 $\pm$       0.00 \\
qbert               &    5512.50 $\pm$     741.08 &   15297.50 $\pm$    1244.47&      40.24 $\pm$       5.58 &     113.86 $\pm$       9.36 \\
riverraid           &    8237.00 $\pm$      97.09 &   11160.00 $\pm$     733.06&      43.72 $\pm$       0.62 &      62.24 $\pm$       4.65 \\
road\_runner        &   28440.00 $\pm$    1215.99 &   51060.00 $\pm$    1560.72&     362.91 $\pm$      15.52 &     651.67 $\pm$      19.92 \\
robotank            &      29.60 $\pm$       2.15 &      46.80 $\pm$       3.42&     282.47 $\pm$      22.22 &     459.79 $\pm$      35.29 \\
seaquest            &    1888.00 $\pm$      63.26 &    9953.00 $\pm$     973.02&       4.33 $\pm$       0.15 &      23.54 $\pm$       2.32 \\
skiing              &  -16244.00 $\pm$     592.28 &  -15438.10 $\pm$    1573.39&       6.69 $\pm$       4.64 &      13.01 $\pm$      12.33 \\
solaris             &    1794.00 $\pm$     279.04 &    2194.00 $\pm$     417.91&       5.03 $\pm$       2.52 &       8.64 $\pm$       3.77 \\
space\_invaders     &     793.50 $\pm$      90.61 &    1771.50 $\pm$     201.95&      42.45 $\pm$       5.96 &     106.76 $\pm$      13.28 \\
star\_gunner        &   44860.00 $\pm$    5157.74 &   60120.00 $\pm$    1953.60&     461.05 $\pm$      53.80 &     620.24 $\pm$      20.38 \\
surround            &       2.50 $\pm$       1.04 &       4.00 $\pm$       0.62&      75.76 $\pm$       6.31 &      84.85 $\pm$       3.74 \\
tennis              &      -0.10 $\pm$       0.09 &      23.10 $\pm$       0.26&     152.90 $\pm$       0.61 &     302.58 $\pm$       1.69 \\
time\_pilot         &   10890.00 $\pm$     787.46 &   22330.00 $\pm$    2443.11&     440.77 $\pm$      47.40 &    1129.42 $\pm$     147.07 \\
tutankham           &     218.50 $\pm$      13.53 &     254.60 $\pm$       9.99&     132.59 $\pm$       8.66 &     155.70 $\pm$       6.40 \\
up\_n\_down         &  175083.00 $\pm$   16341.05 &   82913.00 $\pm$   12142.08&    1564.09 $\pm$     146.43 &     738.18 $\pm$     108.80 \\
venture             &       0.00 $\pm$       0.00 &       0.00 $\pm$       0.00&       0.00 $\pm$       0.00 &       0.00 $\pm$       0.00 \\
video\_pinball      &   59898.40 $\pm$   23875.14 &  198845.20 $\pm$   98768.54&     339.02 $\pm$     135.13 &    1125.46 $\pm$     559.03 \\
wizard\_of\_wor     &    6960.00 $\pm$    1730.97 &    7890.00 $\pm$    1595.77&     152.55 $\pm$      41.28 &     174.73 $\pm$      38.06 \\
yars\_revenge       &   12825.70 $\pm$    2065.90 &   41271.70 $\pm$    4726.72&      18.90 $\pm$       4.01 &      74.16 $\pm$       9.18 \\
zaxxon              &   11520.00 $\pm$     646.81 &   18820.00 $\pm$     754.69&     125.67 $\pm$       7.08 &     205.53 $\pm$       8.26 \\
\midrule
Median & & &     117.56 &     155.70 \\

\bottomrule
\end{tabular}
\end{footnotesize}

\caption{Multi-task Atari-57 scores by level after 11.4B total (200M per level) environment frames. All entries show mean $\pm$ standard deviation. Data for IMPALA (``PopArt-IMPALA'') was obtained from the authors of \citeauthor{Hessel2018}~(\citeyear{Hessel2018}). Human-normalized scores are calculated as $(E - R) / (H - R) \times 100$, where $E$ is the episode reward, $R$ the episode reward obtained by a random agent, and $H$ is the episode reward obtained by a human.}
\label{table:atari57_per_level_scores}

\end{center}
\end{table}


\begin{table*}[t]
\begin{center}
\begin{small}
\begin{tabular}{lcc}
\toprule
{\sc Setting} & {\sc Single-task} & {\sc Multi-task} \\
\midrule
Agent discount & \multicolumn{2}{c}{0.99} \\
Image height & \multicolumn{2}{c}{72} \\
Image width  & \multicolumn{2}{c}{96} \\
Number of action repeats & \multicolumn{2}{c}{4} \\
Number of LSTM layers & 2 & 3 \\
Pixel-control cost & \multicolumn{2}{c}{$2\times10^{-3}$} \\
$T_\text{target}$ & \multicolumn{2}{c}{10} \\
$\epsilon_\eta$ & $0.1$ & $0.5$ \\
$\epsilon_\alpha$ (log-uniform) & $[0.001,\ 0.01)$ & $[0.01,\ 0.1)$ \\
\bottomrule
\end{tabular}

\caption{Settings for DMLab.}
\label{table:dmlab_settings}

\end{small}
\end{center}
\end{table*}


\begin{table*}[t]
\begin{center}
\begin{small}
\begin{tabular}{lcc}
\toprule
{\sc Setting} & {\sc Single-task} & {\sc Multi-task} \\
\midrule
Environment discount on end of life & 1 & 0 \\
Agent discount & 0.997 & 0.99 \\
Clipped reward range & no clipping & $[-1, 1]$ \\
Max episode length & \multicolumn{2}{c}{30 mins (108,000 frames)} \\
Image height & \multicolumn{2}{c}{84} \\
Image width  & \multicolumn{2}{c}{84} \\
Grayscale    & \multicolumn{2}{c}{True} \\
Number of stacked frames & \multicolumn{2}{c}{4} \\
Number of action repeats & \multicolumn{2}{c}{4} \\
TrXL: Key/Value size & 32 & $\cdot$ \\
TrXL: Number of heads & 4 & $\cdot$ \\
TrXL: Number of layers & 8 & $\cdot$ \\
TrXL: MLP size & 512 & $\cdot$ \\
$T_\text{target}$ & 1000 & 100 \\
$\epsilon_\eta$ & \multicolumn{2}{c}{$2\times10^{-2}$} \\
$\epsilon_\alpha$ (log-uniform) & $[0.005, 0.01)$ & $[0.001, 0.01)$ \\
\bottomrule
\end{tabular}

\caption{Settings for Atari. TrXL: Transformer-XL.}
\label{table:atari_settings}

\end{small}
\end{center}
\end{table*}


\begin{table}[t]
\begin{center}
\begin{small}
\begin{tabular}{lccc}
\toprule
{\sc Setting} & {\sc Humanoid-Pixels} & {\sc Humanoid-state} & {\sc OpenAI Gym} \\
\midrule
Agent discount & \multicolumn{3}{c}{0.99} \\
Unroll length & 63 & 63 & 39 \\
Image height & 64 & $\cdot$ & $\cdot$ \\
Image width  & 64 & $\cdot$ & $\cdot$ \\
Target update period & \multicolumn{3}{c}{100} \\
$\epsilon_\eta$ & 0.1 & \multicolumn{2}{c}{0.01} \\
$\epsilon_{\alpha_\mu}$ (log-uniform) & $[0.01,\ 1.0)$ & $[0.05,\ 0.5]$ & $[0.005,\ 0.01]$  \\
$\epsilon_{\alpha_\Sigma}$ (log-uniform) & $[5\times10^{-6},\  5\times10^{-5})$ & $[10^{-5},\  5\times10^{-5})$ & $[5\times10^{-6},\  5\times10^{-5})$ \\
\bottomrule
\end{tabular}

\caption{Settings for continuous control. For the humanoid gaps task from pixels the physics time step was 5~ms and the control time step 30~ms.}
\label{table:continuous_control_settings}

\end{small}
\end{center}
\end{table}

\begin{figure}[h]
\begin{center}
\includegraphics[width=0.8\linewidth]{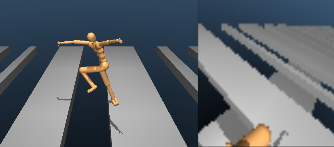}
\caption{Example frame from the humanoid gaps task, with the agent's 64$\times$64 first-person view on the right. The proprioceptive information provided to the agent in addition to the primary pixel observation consisted of joint angles and velocities, root-to-end-effector vectors, root-frame velocity, rotational velocity, root-frame acceleration, and the 3D orientation relative to the $z$-axis.}
\label{fig:humanoid_gaps_frame}
\end{center}
\end{figure}

\begin{figure}[h]
\begin{center}
\includegraphics[width=0.4\linewidth]{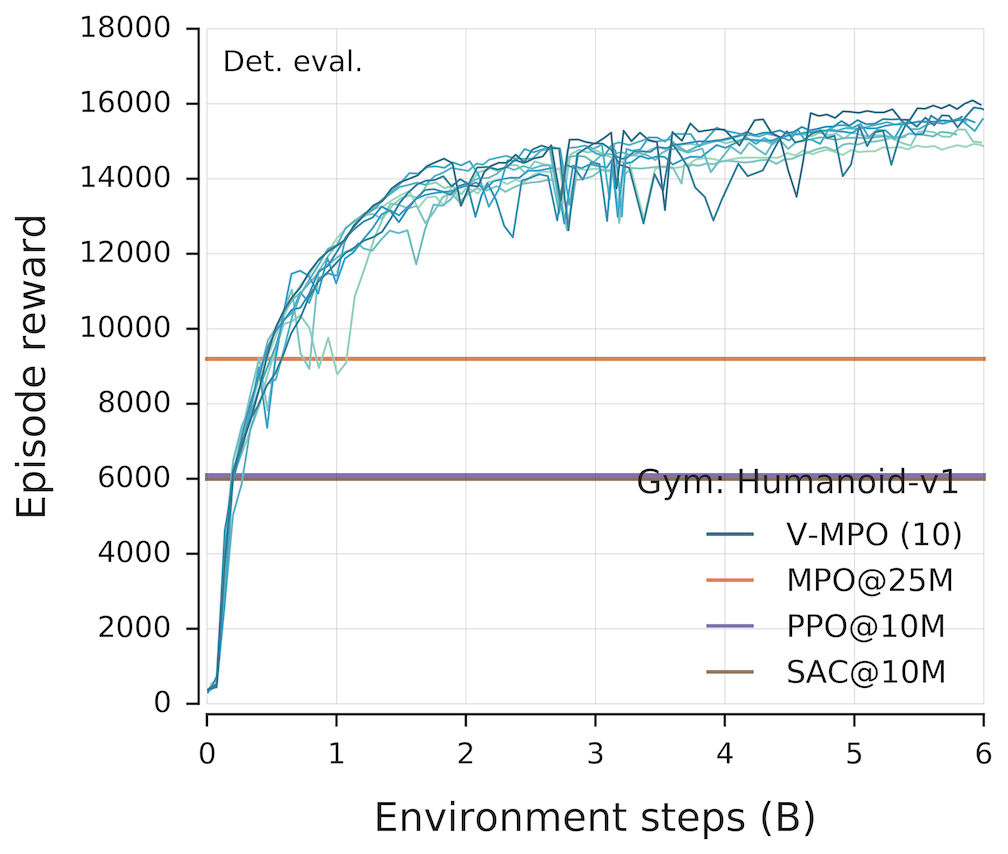}
\caption{17-dimensional Humanoid-V1 task in OpenAI Gym.}
\label{fig:gym_humanoid}
\end{center}
\end{figure}


\end{document}